\documentclass[11pt]{article}
\usepackage[table]{xcolor}
\usepackage[preprint]{acl}
\usepackage{times}
\usepackage{latexsym}
\usepackage[T1]{fontenc}
\usepackage[utf8]{inputenc}
\usepackage{microtype}
\IfFileExists{inconsolata.sty}{\usepackage{inconsolata}}{}
\usepackage{graphicx}
\usepackage{amsmath,amssymb}
\makeatletter
\renewcommand{\maketag@@@}[1]{\hbox{\m@th\normalsize\normalfont#1}}
\makeatother
\usepackage{booktabs,tabularx,array,multirow}
\usepackage{placeins}

\usepackage{tikz}
\usetikzlibrary{arrows.meta,positioning}
\usetikzlibrary{arrows.meta,positioning,svg.path}

\newcommand{\githubicon}{%
  \raisebox{-0.08em}{%
    \tikz[scale=0.42]{%
      \begin{scope}[yscale=-1,yshift=-24pt]
        \fill svg {M12 .297c-6.63 0-12 5.373-12 12 0 5.303 3.438 9.8 8.205 11.385.6 .113 .82-.258 .82-.577 0-.285-.01-1.04-.015-2.04-3.338 .724-4.042-1.61-4.042-1.61-.546-1.387-1.335-1.756-1.335-1.756-1.087-.744 .084-.729 .084-.729 1.205 .084 1.84 1.236 1.84 1.236 1.07 1.835 2.809 1.305 3.495 .998 .108-.776 .417-1.305 .76-1.605-2.665-.3-5.466-1.332-5.466-5.93 0-1.31 .465-2.38 1.235-3.22-.135-.303-.54-1.523 .105-3.176 0 0 1.005-.322 3.3 1.23 .96-.267 1.98-.399 3-.405 1.02 .006 2.04 .138 3 .405 2.28-1.552 3.285-1.23 3.285-1.23 .645 1.653 .24 2.873 .12 3.176 .765 .84 1.23 1.91 1.23 3.22 0 4.61-2.805 5.625-5.475 5.92 .42 .36 .81 1.096 .81 2.22 0 1.606-.015 2.896-.015 3.286 0 .315 .21 .69 .825 .57C20.565 22.092 24 17.592 24 12.297c0-6.627-5.373-12-12-12};%
      \end{scope}%
    }%
  }%
}
\definecolor{prefix3k}{RGB}{255,244,204}
\definecolor{prefix10k}{RGB}{221,235,247}
\definecolor{prefix50k}{RGB}{226,239,218}
\definecolor{prefix100k}{RGB}{244,204,204}
\definecolor{bestscore}{RGB}{247,231,220}
\newcommand{\best}[1]{\cellcolor{bestscore}\textbf{#1}}

\definecolor{traceblue}{RGB}{255,244,204}
\definecolor{tracepink}{RGB}{221,235,247}
\definecolor{tracepurple}{RGB}{226,239,218}
\definecolor{tracegray}{RGB}{244,204,204}

\title{
CAMR: Compatibility-Aware Memory Replay for Continual Learning\\ in Large Language Models
}
\author{
  \normalfont Jingpu Yang\textsuperscript{1,2} \quad
  Tianhao Li\textsuperscript{3} \quad
  Fengxian Ji\textsuperscript{3} \\
  \normalfont Yufeng Wang\textsuperscript{1} \quad
  Shuxin Zheng\textsuperscript{2} \quad
  Xiaoqing Zhang\textsuperscript{2}\thanks{Corresponding author.} \\[0.6em]
  \resizebox{\textwidth}{!}{%
    \normalfont\small
    \textsuperscript{1}Beihang University \quad
    \textsuperscript{2}Beijing Zhongguancun Academy \quad
    \textsuperscript{3}Northeastern University%
  }
}

\newcommand{\method}{\textsc{CAMR}}
\newcommand{\addact}{\textsc{Add}}
\newcommand{\mergeact}{\textsc{Merge}}
\newcommand{\ignoreact}{\textsc{Ignore}}

\begin{document}
\maketitle
\begin{abstract}
Continual learning requires large language models to acquire new knowledge and capabilities while retaining what they have learned.
Fine-tuning and replay support acquisition and retention, but deciding whether incoming examples need supervision and which historical examples to revisit remains a challenge.
We propose Compatibility-Aware Memory Replay (\method{}), an adaptive framework that jointly controls current-example supervision and historical replay.
\method{} combines compatibility checks with a learned memory gate to guide an online policy over three actions: learning incoming examples with general replay, jointly reinforcing compatible related examples, and skipping direct supervision of verified mastered duplicates.
The policy learns from acquisition, retention, and training-cost feedback under a bounded per-arrival replay budget.
We evaluate \method{} with Qwen3-8B-Base on continual factual updating and cross-task and cross-domain learning.
On WikiBigEdit, \method{} consistently leads evaluated methods in Original Match from 3K to 100K facts, reaching 93.66\% after just 3K facts and surpassing Full-history Replay (LoRA) after 100K facts.
It also achieves the highest final scores on all four selected TRACE tasks and maintains or improves earlier-task performance relative to post-acquisition levels in a sequence spanning financial question answering, mathematical reasoning, code generation, and summarization. 
\githubicon\ \href{https://anonymous.4open.science/r/CAMR-FE92/}{Github.}
\end{abstract}

\section{Introduction}
\label{sec:introduction}
Continual learning enables large language models (LLMs) to acquire new knowledge and capabilities while retaining prior knowledge, allowing them to adapt as knowledge and task requirements evolve.
However, successive updates can disrupt previously learned knowledge, giving rise to the stability--plasticity problem \citep{mccloskey1989catastrophic,delange2022survey}.

\begin{figure*}[!t]
\centering
\includegraphics[width=\textwidth]{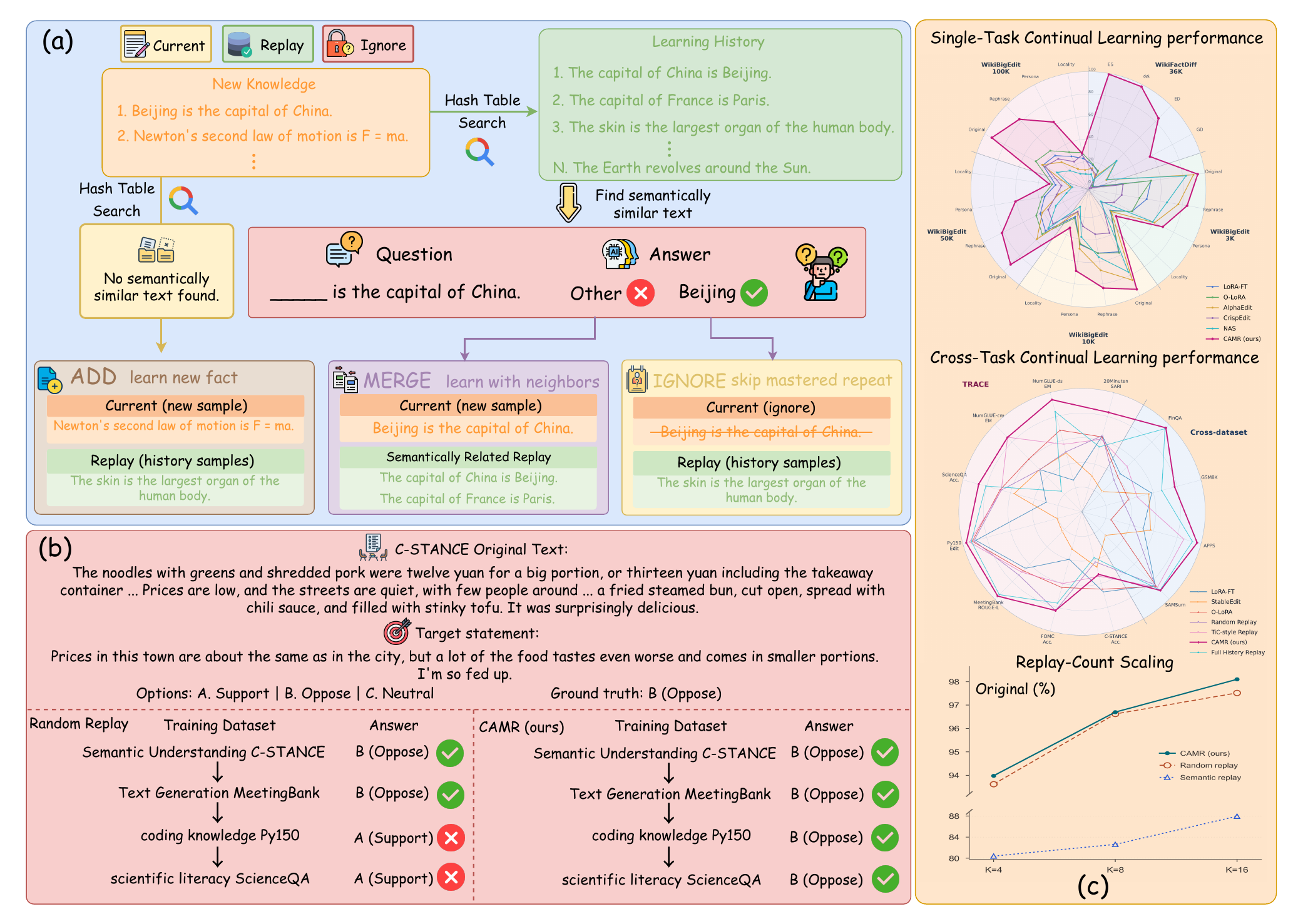}
\caption{Overview of \method{} and its empirical evaluation. \textbf{(a)} Schematic illustration of \method{}. It shows how \addact{}, \mergeact{}, and \ignoreact{} handle incoming data during continual learning. \textbf{(b)} Example comparison of \method{} with Random Replay in the multi-task setting. \textbf{(c)} Results for single-task and cross-task settings, together with Replay-Count Scaling experiments.}
\label{fig:intro}
\end{figure*}

One line of work mitigates interference by controlling parameter updates.
Parameter-space methods constrain update directions or separate task-specific subspaces \citep{olora2023}, while knowledge-editing methods make targeted changes to factual associations, with subsequent advances improving stability over successive edits \citep{meng2022locating,meng2023mass,alphaedit2025,nas2026}.
These approaches improve learning under given supervision, but their update mechanisms do not directly determine which incoming and historical examples should be used for training.

Experience replay complements these approaches by revisiting historical examples to support retention \citep{rolnick2019experience}.
Full-history replay incurs growing training costs, while random replay does not account for individual learning needs.
Adaptive methods select historical examples based on relevance, interference risk, or transfer value, as illustrated by MIR \citep{aljundi2019online} and TSR \citep{meng2026rethinking}.
Yet selecting historical examples addresses only part of supervision allocation.
Incoming examples may require further learning, benefit from joint training with compatible related history, or need no further direct supervision when they repeat material the model has already mastered.
Meanwhile, unrelated historical examples still need replay coverage to support retention.
These considerations motivate coordinating current-example supervision with the composition of historical replay.

We propose \emph{Compatibility-Aware Memory Replay} (\method{}), which jointly controls current-example supervision and historical replay under a bounded per-arrival replay budget (Figure~\ref{fig:intro}).
Compatibility checks identify eligible history, and a learned memory gate summarizes retrieved candidates.
The resulting memory representation and model mastery guide an online policy over three actions:
\addact{} trains on the incoming example with general replay;
\mergeact{} additionally prioritizes compatible related history within the replay budget;
and \ignoreact{} skips direct supervision of a verified mastered duplicate while allowing historical replay.
The policy adapts online through feedback on acquisition, retention, and training cost.
The language model undergoes full-parameter supervised fine-tuning, with retrieval and policy decisions confined to training.

We evaluate \method{} with Qwen3-8B-Base on WikiBigEdit, WikiFactDiff, a four-task TRACE sequence, and a cross-domain sequence.
On WikiBigEdit, \method{} consistently achieves the highest Original Match among evaluated methods at checkpoints from 3K to 100K facts.
It reaches 93.66\% after just 3K facts, surpassing Full-history Replay (LoRA) after 100K facts (92.07\%).
On WikiFactDiff, it improves final-stage generalization success over Random Replay (Full FT) by 1.25 percentage points.
It also achieves the highest final scores on all four selected TRACE tasks and maintains or improves earlier-task performance relative to post-acquisition levels in the FinQA $\rightarrow$ GSM8K $\rightarrow$ APPS $\rightarrow$ SAMSum sequence.
Replay-count scaling and ablations examine sensitivity to replay capacity and the contributions of key policy components.
Our contributions are threefold:
\begin{itemize}
    \item We introduce a compatibility-aware framework that jointly determines current-fact supervision and historical replay under a bounded per-arrival replay budget.
    \item We develop an online policy that combines model mastery and gated historical information to adapt supervision decisions through learning feedback.
    \item We demonstrate the effectiveness of \method{} in continual factual updating and cross-task and cross-domain continual learning, supported by replay-count scaling and component ablations.
\end{itemize}

\section{Related Work}
\label{sec:related}
\paragraph{Continual learning and parameter updates.}
Continual-learning methods mitigate forgetting by constraining parameter updates, through regularization \citep{kirkpatrick2017ewc}, gradient projection \citep{wang2021adamnscl}, or orthogonal low-rank subspaces in LLMs \citep{olora2023}.
Knowledge-editing methods rewrite specific factual associations \citep{meng2022locating,meng2023mass}, and later work improves stability over long edit sequences \citep{alphaedit2025}.
These methods control how parameters change, whereas \method{} decides which examples supervise each update.

\paragraph{Experience replay and adaptive selection.}
Experience replay mixes stored examples with new data to support retention \citep{rolnick2019experience}.
Adaptive variants select history by interference risk \citep{aljundi2019online} or transfer value \citep{meng2026rethinking}.
These methods choose which history to revisit, while \method{} also decides how the incoming example itself is supervised.

\paragraph{Memory mechanisms and learned supervision policies.}
Reinforcement learning has been used to learn which data a model should train on, for example through teacher models that select examples \citep{fan2018learning} and bandit-based curricula driven by learning progress \citep{graves2017automated}.
Our writer follows this direction with an actor--critic policy \citep{mnih2016asynchronous} rewarded by acquisition, retention, and training cost.
Its state includes a gated summary of history, adapted from the conditional memory of Engram \citep{cheng2026conditional}.

\begin{figure*}[t]
\centering
\includegraphics[width=\textwidth]{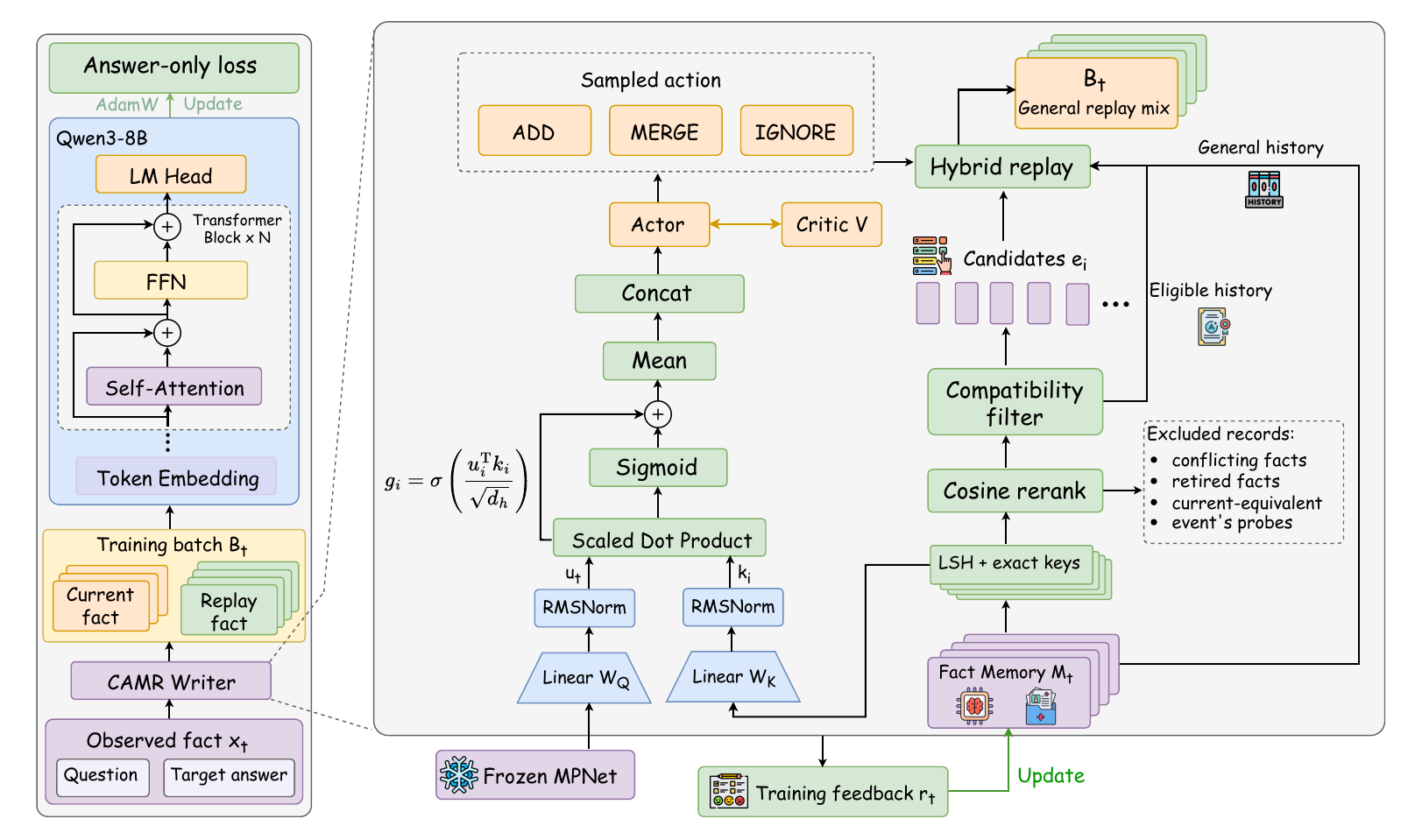}
\caption{Overview of \method{}. The writer retrieves and filters compatible history, then selects \addact{}, \mergeact{}, or \ignoreact{} to form the training batch; feedback from the updated language model trains the writer.}
\label{fig:overview}
\end{figure*}

\section{Problem Setting}
\label{sec:formulation}

We consider a stream of supervised examples
$\mathcal{S}=(x_1,\ldots,x_T)$, where $x_t=(q_t,y_t,m_t)$
contains an input, a target output, and optional identity,
task, or version metadata.
At arrival $t$, the learner may use the current and previously
observed examples, but has no access to future arrivals or
held-out targets.
The objective is to acquire new knowledge and capabilities
while retaining prior learning that remains valid.
For factual updates, supervision is trusted: an authoritative
replacement makes the new answer valid and retires the old one.



Let $p_{\theta_t}$ and $\mathcal{M}_t$ denote the model and
external memory before arrival $t$.
The writer jointly controls current-example supervision and
historical replay under a per-arrival replay cap $K$, which counts
historical replay examples and excludes the current example:
\addact{} trains on the current example with general replay;
\mergeact{} additionally prioritizes compatible related history
within the same budget;
and \ignoreact{} skips direct supervision of a verified mastered
duplicate while allowing history-only replay.
All replay examples must satisfy the compatibility criteria.
Training updates the model to $p_{\theta_{t+1}}$.
Regardless of the action, the observed example is stored or
linked to an equivalent record in $\mathcal{M}_{t+1}$.
Retired or conflicting records are archived but excluded from
replay and retention evaluation.
Evaluation is closed-book, with retrieval and writer decisions
confined to training.


\section{Method}
\label{sec:method}
Figure~\ref{fig:overview} summarizes \method{}.
The writer combines model mastery with a gated representation
of compatible history to select \addact{}, \mergeact{}, or
\ignoreact{}, jointly determining current-example supervision
and historical replay.
The resulting nonempty batches update the language model
using answer-only loss.
Pre- and post-event measurements provide acquisition,
retention, and training-cost feedback for writer updates
at rollout boundaries.
The language model and writer use separate optimizers,
while the sentence encoder remains frozen.




\subsection{Memory Retrieval and Compatibility}
\label{sec:memory}

Memory stores observed examples with their embeddings,
identity/version metadata, and replay statistics.
For each incoming example $x_t$, we distinguish equivalent,
conflicting, and retired historical records.
With structured metadata, a slot is defined by
(subject, relation, scope), and fact identity additionally
includes object and target.
Records with the same fact identity are equivalent;
different values within the same slot are treated as conflicts.
Without structured metadata, input--output pairs that match after
conservative Unicode and whitespace normalization are equivalent,
while different targets for the same normalized input are conflicting.
A record is retired only by an explicit replacement reference
or an authoritative update to a declared single-valued slot.
Conflicting records are excluded from the current event's
replay but are not automatically retired.
We define eligible history $\mathcal{H}_t$ by excluding
retired, current-equivalent, and conflicting records from memory.
Equivalent records remain available to determine whether
\ignoreact{} is permitted.

Before action selection, we uniformly sample a probe set
$\mathcal{P}_t$ from $\mathcal{H}_t$ without replacement, with
$|\mathcal{P}_t|=\min(P,\lfloor|\mathcal{H}_t|/2\rfloor)$.
These probes measure historical loss changes across the
event and are excluded from its training batch.
The remaining records form the replay pool
$\mathcal{U}_t=\mathcal{H}_t\setminus\mathcal{P}_t$.

As shown in Figure~\ref{fig:overview}, retrieval combines
semantic search with exact lookups.
A frozen MPNet encoder \citep{song2020mpnet} maps each
input--output pair to a mean-pooled, L2-normalized embedding.
Let $e_t$ and $e_i$ denote current and historical embeddings.
Random-hyperplane locality-sensitive hashing
\citep{charikar2002similarity} retrieves semantically similar
records, while exact indices retrieve records by fact
identity and slot.
The results are reranked by cosine similarity to form
a capped shortlist.
We then apply the compatibility rules above and exclude
probe records.
The remaining records with similarity above $\tau_s$
form the candidate set $\mathcal{C}_t\subseteq\mathcal{U}_t$.
These candidates feed the writer's relevance gate,
while $\mathcal{U}_t$ supplies general historical replay.
After action selection, the action-specific replay rule selects
$\mathcal{R}_t\subseteq\mathcal{U}_t$ subject to
$|\mathcal{R}_t|\leq K$.
After the event, new examples are stored in memory,
while equivalent arrivals are linked to existing records.

\subsection{Writer Policy and Replay}
\label{sec:policy}

Given the compatible candidates $\mathcal{C}_t$,
the writer summarizes related history, selects a supervision
action, and constructs the training batch
(Figure~\ref{fig:overview}).

We adapt the sigmoid query--key gating mechanism of Engram
\citep{cheng2026conditional} to current and historical
example embeddings.
The frozen MPNet embeddings are projected into a shared
$d_h$-dimensional space and normalized using RMSNorm
\citep{zhang2019rmsnorm}:
\begin{equation}
\resizebox{0.82\linewidth}{!}{$\displaystyle
u_t=\operatorname{RMSNorm}(W_Qe_t),\quad
k_i=\operatorname{RMSNorm}(W_Ke_i).$}
\end{equation}
The gate assigns each candidate a relevance weight:
\begin{equation}
g_{ti}=\sigma\!\left(
\frac{u_t^\top k_i}{\sqrt{d_h}}
\right).
\label{eq:camr_gate}
\end{equation}
We aggregate the gated candidate representations and
their mean gate value as
\begin{equation}
v_t=\frac{1}{|\mathcal{C}_t|}
\sum_{i\in\mathcal{C}_t}g_{ti}k_i,\qquad
\bar g_t=\frac{1}{|\mathcal{C}_t|}
\sum_{i\in\mathcal{C}_t}g_{ti}.
\label{eq:camr_summary}
\end{equation}
Both are set to zero when $\mathcal{C}_t$ is empty.
The gate operates in the writer, outside the language model.

The actor receives the concatenated state
$z_t=[f_t;u_t;v_t;\bar g_t]$.
The eight features in $f_t$ encode top similarity,
equivalence, current NLL, probe NLL, ambiguity,
candidate count, stream progress, and update frequency.
Two actor heads produce masked probabilities
$\rho_t^I=b_t^I\sigma(a_I(z_t))$ and
$\rho_t^M=b_t^M\sigma(a_M(z_t))$, where
$b_t^I,b_t^M\in\{0,1\}$ indicate action eligibility.
The action distribution is
\begin{equation}
\resizebox{0.82\linewidth}{!}{$\displaystyle
\begin{gathered}[t]
\pi_\phi(\ignoreact{} \mid z_t)=\rho_t^I,\quad
\pi_\phi(\mergeact{} \mid z_t)=(1-\rho_t^I)\rho_t^M, \\
\pi_\phi(\addact{} \mid z_t)=(1-\rho_t^I)(1-\rho_t^M).
\end{gathered}$}
\label{eq:actions}
\end{equation}
\mergeact{} is eligible only when $\mathcal{C}_t$ is nonempty.
\ignoreact{} requires an equivalent active record,
current NLL $\leq\tau_m$, no retirement triggered by
the current arrival, and a fresh greedy answer matching
the full target after conservative Unicode and
whitespace normalization.
If neither action is eligible, the policy selects \addact{}.
The critic estimates the state value for online learning,
as described in Section~\ref{sec:lm_update}.

The sampled action determines current-example supervision
and the composition of historical replay.
\addact{} and \ignoreact{} sample uniformly from
the replay pool $\mathcal{U}_t$.
\mergeact{} combines general history with related candidates:
it reserves part of the replay budget for uniform sampling
and allocates the remainder to compatible candidates
from $\mathcal{C}_t$.
Unused candidate slots are filled by uniform sampling
from the remaining replay pool.
All replay sampling is without replacement and selects
at most $K$ records.
The training batch contains the current example and
$\mathcal{R}_t$ under \addact{} or \mergeact{},
and only $\mathcal{R}_t$ under \ignoreact{}.
Each replayed example remains a separate input--output pair.

\begin{table*}[!t]
\centering
\scriptsize
\setlength{\tabcolsep}{2.0pt}
\renewcommand{\arraystretch}{1.12}
\definecolor{scoregain}{RGB}{34,139,105}
\definecolor{scoreloss}{RGB}{194,74,66}
\definecolor{scoreflat}{RGB}{120,120,120}
\definecolor{camrrow}{RGB}{235,243,253}
\newcommand{\scorechange}[2]{\,\scalebox{0.78}{\textcolor{#1}{(#2)}}}
\caption{
Long-stream match accuracy (\%) and checkpoint change.
All checkpoints belong to the same 100K continual stream.
Parentheses at 100K report the change from 3K in percentage points.
The light peach shading
\protect\colorbox{bestscore}{\phantom{\rule{1.15ex}{1.15ex}}}
marks the highest value in each column, these values are also bold.
The additional \method{} LoRA run uses $K=8$ and a learning rate of
$2\times10^{-4}$ through 50K updates and $10^{-4}$ thereafter,
continuing the same model and optimizer.
}
\label{tab:prefix-comparison}

\resizebox{\textwidth}{!}{%
\begin{tabular}{@{}l*{16}{r}@{}}
\toprule
\multirow{2}{*}{Method}
& \multicolumn{4}{c}{\cellcolor{prefix3k}3K}
& \multicolumn{4}{c}{\cellcolor{prefix10k}10K}
& \multicolumn{4}{c}{\cellcolor{prefix50k}50K}
& \multicolumn{4}{c}{\cellcolor{prefix100k}100K (change vs.\ 3K)} \\
& \multicolumn{1}{>{\columncolor{prefix3k}}c}{Original}
& \multicolumn{1}{>{\columncolor{prefix3k}}c}{Rephrase}
& \multicolumn{1}{>{\columncolor{prefix3k}}c}{Persona}
& \multicolumn{1}{>{\columncolor{prefix3k}}c}{Locality}
& \multicolumn{1}{>{\columncolor{prefix10k}}c}{Original}
& \multicolumn{1}{>{\columncolor{prefix10k}}c}{Rephrase}
& \multicolumn{1}{>{\columncolor{prefix10k}}c}{Persona}
& \multicolumn{1}{>{\columncolor{prefix10k}}c}{Locality}
& \multicolumn{1}{>{\columncolor{prefix50k}}c}{Original}
& \multicolumn{1}{>{\columncolor{prefix50k}}c}{Rephrase}
& \multicolumn{1}{>{\columncolor{prefix50k}}c}{Persona}
& \multicolumn{1}{>{\columncolor{prefix50k}}c}{Locality}
& \multicolumn{1}{>{\columncolor{prefix100k}}c}{Original}
& \multicolumn{1}{>{\columncolor{prefix100k}}c}{Rephrase}
& \multicolumn{1}{>{\columncolor{prefix100k}}c}{Persona}
& \multicolumn{1}{>{\columncolor{prefix100k}}c}{Locality} \\
\midrule
\multicolumn{17}{c}{\textit{Unmodified base models}} \\
\midrule
Qwen3-8B-Base
& 14.67 & 13.71 & 11.20 & 5.88 & 13.16 & 12.51 & 9.33 & 9.62
& 12.04 & 11.90 & 8.76 & 10.55 & 12.88\scorechange{scoreloss}{-1.79} & 12.33\scorechange{scoreloss}{-1.38} & 9.08\scorechange{scoreloss}{-2.12} & 13.09\scorechange{scoregain}{+7.21} \\
Qwen3-14B-Base
& 21.63 & 20.20 & 11.53 & 29.41 & 20.42 & 19.69 & 9.99 & 21.15
& 17.26 & 16.55 & 8.72 & 17.58 & 17.40\scorechange{scoreloss}{-4.23} & 16.76\scorechange{scoreloss}{-3.44} & 8.74\scorechange{scoreloss}{-2.79} & 18.75\scorechange{scoreloss}{-10.66} \\
\midrule
\multicolumn{17}{c}{\textit{Fine-tuning and parameter-space methods}} \\
\midrule
Fine-Tuning (LoRA)
& 51.40 & 47.83 & 38.57 & 21.97 & 56.83 & 52.34 & 44.61 & 21.95
& 48.27 & 43.10 & 34.70 & 24.62 & 41.08\scorechange{scoreloss}{-10.32} & 37.40\scorechange{scoreloss}{-10.43} & 28.23\scorechange{scoreloss}{-10.34} & 23.21\scorechange{scoregain}{+1.24} \\
Sequential Fine-Tuning (Full FT)
& 60.73 & 56.00 & 47.63 & 18.44 & 62.30 & 57.64 & 51.16 & 21.18
& 52.79 & 48.03 & 38.98 & 21.42 & 45.83\scorechange{scoreloss}{-14.90} & 41.47\scorechange{scoreloss}{-14.53} & 34.36\scorechange{scoreloss}{-13.27} & 22.66\scorechange{scoregain}{+4.22} \\
Joint Fine-Tuning (Full FT)
& 61.20 & 56.10 & 48.47 & 19.60 & 63.12 & 58.54 & 51.72 & 21.59
& 58.19 & 52.95 & 44.09 & 23.13 & 56.36\scorechange{scoreloss}{-4.84} & 51.08\scorechange{scoreloss}{-5.02} & 42.15\scorechange{scoreloss}{-6.32} & 24.32\scorechange{scoregain}{+4.72} \\
O-LoRA
& 51.63 & 47.27 & 38.53 & 22.68 & 57.84 & 52.55 & 43.86 & 23.44
& 52.28 & 47.12 & 37.12 & 27.64 & 48.31\scorechange{scoreloss}{-3.32} & 43.45\scorechange{scoreloss}{-3.82} & 32.33\scorechange{scoreloss}{-6.20} & 28.35\scorechange{scoregain}{+5.67} \\
Adam-NSCL
& 15.77 & 14.72 & 11.80 & 11.76 & 17.93 & 16.49 & 11.60 & 19.23
& 17.43 & 16.69 & 11.70 & 16.41 & 18.09\scorechange{scoregain}{+2.32} & 17.29\scorechange{scoregain}{+2.57} & 12.19\scorechange{scoregain}{+0.39} & 17.58\scorechange{scoregain}{+5.82} \\
LoRA-Merge
& 25.98 & 24.78 & 20.14 & 29.41 & 26.09 & 26.28 & 19.75 & 19.23
& 31.11 & 27.54 & 21.14 & 27.34 & 35.31\scorechange{scoregain}{+9.33} & 30.80\scorechange{scoregain}{+6.02} & 22.44\scorechange{scoregain}{+2.30} & \best{29.10}\scorechange{scoreloss}{-0.31} \\
\midrule
\multicolumn{17}{c}{\textit{Knowledge-editing methods}} \\
\midrule
LocFT
& 29.03 & 26.30 & 21.73 & 18.36 & 40.49 & 36.08 & 30.24 & 21.68
& 41.98 & 34.22 & 25.89 & 25.98 & 41.98\scorechange{scoregain}{+12.95} & 34.02\scorechange{scoregain}{+7.72} & 23.81\scorechange{scoregain}{+2.08} & 27.34\scorechange{scoregain}{+8.98} \\
ROME
& 37.07 & 35.47 & 27.33 & 13.77 & 25.72 & 24.46 & 17.99 & 13.55
& 17.82 & 17.29 & 12.72 & 14.28 & 17.94\scorechange{scoreloss}{-19.13} & 16.98\scorechange{scoreloss}{-18.49} & 12.37\scorechange{scoreloss}{-14.96} & 14.41\scorechange{scoregain}{+0.64} \\
R-ROME
& 37.30 & 35.17 & 28.50 & 13.67 & 26.63 & 25.52 & 20.12 & 13.88
& 19.55 & 17.80 & 13.22 & 14.90 & 19.52\scorechange{scoreloss}{-17.78} & 17.99\scorechange{scoreloss}{-17.18} & 13.35\scorechange{scoreloss}{-15.15} & 15.52\scorechange{scoregain}{+1.85} \\
AlphaEdit
& 89.76 & 75.42 & 61.05 & 23.53 & 84.82 & 67.84 & 48.43 & 17.31
& 58.95 & 41.09 & 22.18 & 11.33 & 41.72\scorechange{scoreloss}{-48.04} & 29.20\scorechange{scoreloss}{-46.22} & 15.54\scorechange{scoreloss}{-45.51} & 12.70\scorechange{scoreloss}{-10.83} \\
NAS
& 83.07 & 66.10 & 59.23 & 10.28 & 76.96 & 55.90 & 37.66 & 11.97
& 44.75 & 27.45 & 14.94 & 9.64 & 27.08\scorechange{scoreloss}{-55.99} & 18.51\scorechange{scoreloss}{-47.59} & 10.74\scorechange{scoreloss}{-48.49} & 8.95\scorechange{scoreloss}{-1.33} \\
CrispEdit
& 26.10 & 25.17 & 20.77 & 11.52 & 38.11 & 35.36 & 28.19 & 16.99
& 38.89 & 35.19 & 23.88 & 19.34 & 38.73\scorechange{scoregain}{+12.63} & 34.35\scorechange{scoregain}{+9.18} & 23.34\scorechange{scoregain}{+2.57} & 18.16\scorechange{scoregain}{+6.64} \\
UltraEdit
& 17.14 & 15.72 & 12.50 & 5.88 & 15.79 & 14.85 & 10.85 & 9.62
& 15.80 & 15.20 & 11.07 & 13.28 & 19.11\scorechange{scoregain}{+1.97} & 17.51\scorechange{scoregain}{+1.79} & 12.66\scorechange{scoregain}{+0.16} & 17.58\scorechange{scoregain}{+11.70} \\
StableEdit
& 24.13 & 22.80 & 16.07 & 18.00 & 27.26 & 24.91 & 17.63 & 20.04
& 35.49 & 28.89 & 16.48 & 25.59 & 35.43\scorechange{scoregain}{+11.30} & 28.84\scorechange{scoregain}{+6.04} & 17.36\scorechange{scoregain}{+1.29} & 24.10\scorechange{scoregain}{+6.10} \\
WISE-Retrieve
& 23.77 & 21.91 & 17.31 & 17.65 & 23.80 & 21.39 & 15.60 & 23.08
& 13.76 & 12.86 & 11.09 & 13.67 & 11.37\scorechange{scoreloss}{-12.40} & 10.54\scorechange{scoreloss}{-11.37} & 10.08\scorechange{scoreloss}{-7.23} & 12.89\scorechange{scoreloss}{-4.76} \\
\midrule
\multicolumn{17}{c}{\textit{Replay-based continual-learning methods}} \\
\midrule
Random Replay (Full FT)
& 87.96 & 80.23 & 67.46 & 23.53 & 91.32 & \best{84.75} & \best{68.94} & 23.92
& 90.74 & \best{81.62} & 60.52 & 25.34 & 89.12\scorechange{scoregain}{+1.16} & 80.89\scorechange{scoregain}{+0.66} & 61.99\scorechange{scoreloss}{-5.47} & 24.73\scorechange{scoregain}{+1.20} \\
Random Replay (LoRA)
& 68.66 & 62.07 & 51.98 & 23.53 & 64.34 & 59.53 & 50.28 & 19.23
& 47.73 & 45.06 & 35.24 & 21.88 & 41.63\scorechange{scoreloss}{-27.03} & 38.68\scorechange{scoreloss}{-23.39} & 29.16\scorechange{scoreloss}{-22.82} & 21.48\scorechange{scoreloss}{-2.05} \\
Full-history Replay (LoRA)
& 50.92 & 46.02 & 36.61 & 29.41 & 87.37 & 75.24 & 56.47 & 28.85
& 88.86 & 77.10 & 53.38 & 26.95 & 92.07\scorechange{scoregain}{+41.15} & \best{83.53}\scorechange{scoregain}{+37.51} & 61.55\scorechange{scoregain}{+24.94} & 22.71\scorechange{scoreloss}{-6.70} \\
Semantic Replay (LoRA)
& 51.40 & 47.83 & 38.57 & 23.94 & 57.51 & 53.30 & 45.42 & 25.86
& 51.34 & 46.90 & 36.74 & 25.47 & 46.83\scorechange{scoreloss}{-4.57} & 42.51\scorechange{scoreloss}{-5.32} & 33.68\scorechange{scoreloss}{-4.89} & 23.42\scorechange{scoreloss}{-0.52} \\
Relation Oracle Replay (LoRA)
& 68.02 & 61.71 & 50.92 & 29.41 & 63.65 & 58.89 & 47.94 & 19.23
& 47.23 & 44.02 & 34.55 & 28.91 & 41.39\scorechange{scoreloss}{-26.63} & 38.32\scorechange{scoreloss}{-23.39} & 30.72\scorechange{scoreloss}{-20.20} & 23.44\scorechange{scoreloss}{-5.97} \\
TiC-style Replay (LoRA)
& 65.29 & 59.57 & 48.52 & 23.53 & 63.69 & 58.10 & 47.51 & 25.00
& 51.13 & 47.80 & 37.66 & 28.13 & 46.84\scorechange{scoreloss}{-18.45} & 43.92\scorechange{scoreloss}{-15.65} & 34.59\scorechange{scoreloss}{-13.93} & 26.37\scorechange{scoregain}{+2.84} \\
Maximally Interfered Retrieval (LoRA)
& 31.67 & 28.27 & 23.63 & 20.90 & 44.80 & 40.51 & 33.54 & 21.88
& 39.95 & 34.10 & 24.35 & 26.76 & 38.73\scorechange{scoregain}{+7.06} & 33.37\scorechange{scoregain}{+5.10} & 22.90\scorechange{scoreloss}{-0.73} & 26.37\scorechange{scoregain}{+5.47} \\
Transfer Selective Replay (LoRA)
& 46.87 & 41.77 & 34.03 & 25.21 & 49.11 & 44.47 & 36.98 & 26.59
& 44.98 & 39.05 & 30.53 & 27.96 & 45.62\scorechange{scoreloss}{-1.25} & 40.01\scorechange{scoreloss}{-1.76} & 30.22\scorechange{scoreloss}{-3.81} & 28.98\scorechange{scoregain}{+3.77} \\
\rowcolor{camrrow}
\textbf{\method{} (Full FT, ours)}
& \best{93.66} & 84.38 & 69.26 & \best{30.47} & \best{93.86} & 83.84 & 68.38 & \best{33.08}
& \best{91.43} & 81.45 & \best{61.08} & \best{30.47} & \best{92.94}\scorechange{scoreloss}{-0.72} & 82.82\scorechange{scoreloss}{-1.56} & \best{62.61}\scorechange{scoreloss}{-6.65} & 27.73\scorechange{scoreloss}{-2.74} \\
\rowcolor{camrrow}
\textbf{\method{} (LoRA, ours)}
& 92.36 & \best{86.45} & \best{73.06} & 17.65 & 89.34 & 83.42 & 66.99 & 23.08
& 74.18 & 65.32 & 45.32 & 19.14 & 82.49\scorechange{scoreloss}{-9.88} & 71.06\scorechange{scoreloss}{-15.39} & 46.06\scorechange{scoreloss}{-27.00} & 24.41\scorechange{scoregain}{+6.77} \\
\bottomrule
\end{tabular}%
}
\end{table*}

\subsection{Learning Objectives and Online Updates}
\label{sec:lm_update}

We optimize an answer-prediction objective for the language
model and an actor--critic objective for the writer.
Let $\ell_\theta(x)$ denote the average answer-token NLL
over a fact's original question and optional same-answer
training view, excluding evaluation reformulations.
The loss includes end-of-sequence tokens but excludes prompt tokens.
When current-fact supervision is active and replay
is nonempty, the language-model objective is
\begin{equation}
\mathcal{L}^{\mathrm{LM}}_t
=
\alpha \ell_\theta(x_t)
+
\frac{1-\alpha}{|\mathcal{R}_t|}
\sum_{x\in\mathcal{R}_t}\ell_\theta(x),
\label{eq:lm_loss}
\end{equation}
where $\alpha$ balances current and historical supervision.
Current-only updates assign unit weight to the current loss;
history-only updates average replay losses.
An empty \ignoreact{} batch triggers no model update.
We use AdamW \citep{loshchilov2019adamw}, retaining
optimizer state across events.

The writer receives feedback from NLL measurements
before and after each event on the original current
question and the same probes $\mathcal{P}_t$.
Let $\Delta_t^{\mathrm{cur}}$ and
$\Delta_t^{\mathrm{hist}}$ denote current and mean historical
loss reductions, with zero historical gain for empty probes.
The reward balances acquisition, retention, and training cost:
\begin{equation}
r_t
=
\Delta_t^{\mathrm{cur}}
+
\lambda_h\Delta_t^{\mathrm{hist}}
-
\lambda_c\frac{C_t}{C_{\mathrm{ref}}},
\label{eq:reward}
\end{equation}
where $\lambda_h$ and $\lambda_c$ weight retention and cost.
Here $C_t$ counts logical nonpadding training tokens,
including views and repeated steps, and $C_{\mathrm{ref}}$ is a
dataset-specific cost-normalization hyperparameter.
Probe and mastery-check forward passes are excluded
from $C_t$.

Writer parameters remain fixed within each configured
rollout $\mathcal{B}$ and are updated at its end.
The critic predicts $V_t=V_\psi(\operatorname{sg}(z_t))$,
where $\operatorname{sg}$ stops gradients.
Discounted returns and advantages are
\begin{equation}
\begin{gathered}
G_t=r_t+\gamma G_{t+1},\qquad
\bar
A_t=G_t-\operatorname{sg}(V_t).
\end{gathered}
\label{eq:writer_returns}
\end{equation}
At the rollout boundary, the recursion uses the next
observed state's detached value, or zero at stream termination.
All returns are fixed targets.

The actor adjusts action probabilities using advantages,
while the critic fits the returns.
For choice events $\mathcal{D}\subseteq\mathcal{B}$
and policy entropy $H_t$, we minimize
{\small
\begin{equation}
\begin{aligned}
\mathcal{L}_{\pi}
&=
-\frac{1}{|\mathcal{D}|}
\sum_{t\in\mathcal{D}}
\left[
A_t\log\pi_\phi(a_t\mid z_t)+\beta H_t
\right],\\
\mathcal{L}_{V}
&=
\frac{1}{|\mathcal{B}|}
\sum_{t\in\mathcal{B}}(V_t-G_t)^2,
\;
\mathcal{L}_{W}
=
\mathcal{L}_{\pi}+c_v\mathcal{L}_{V}.
\end{aligned}
\label{eq:writer_loss}
\end{equation}
}
Here $\gamma$ controls discounting, $\beta$ weights entropy
regularization, and $c_v$ weights the critic loss.
We set $\mathcal{L}_{\pi}=0$ when $\mathcal{D}$ is empty.
Forced-action events train only the critic;
the relevance gate learns only through the actor.
The complete set of hyperparameters introduced in the \method{}
formulation is reported in
Appendix~\ref{app:objective-hyperparameters}.


\section{Experiments and Discussion}
\label{sec:experiments}

\subsection{Experimental Setup}
We evaluate Qwen3-8B-Base in the closed-book setting, with Qwen3-14B-Base as an unmodified WikiBigEdit reference \citep{qwen3_2025}, on WikiBigEdit \citep{thede2025wikibigedit}, WikiFactDiff \citep{ammar-khodja-etal-2024-wikifactdiff-large}, TRACE (C-STANCE, MeetingBank, Py150, and text-only ScienceQA) \citep{trace2023,cstance2023,meetingbank2023,py1502016,scienceqa2022}, and FinQA $\rightarrow$ GSM8K $\rightarrow$ APPS introductory $\rightarrow$ SAMSum \citep{finqa2021,gsm8k2021,apps2021,samsum2019}; ablations use FOMC, NumGLUE-cm/ds, and Py150 \citep{fomc2023,numglue2022}.
Baselines span unmodified models; fine-tuning and parameter-space methods \citep{lora2022,olora2023,wang2021adamnscl}; knowledge-editing methods \citep{locft2026,meng2022locating,meng2023mass,gupta2024rrome,alphaedit2025,wise2024,nas2026,crispedit2026,stableedit2026,ultraedit2026}; and replay-based methods \citep{rolnick2019experience,aljundi2019online,meng2026rethinking,ticlm2025}.
We report Match/Exact, ES/GS, ED/GD, accuracy, ROUGE-L, edit similarity, and pass@1 as appropriate.

\subsection{Single-Task Data Scaling}
We process 100,000 WikiBigEdit facts \citep{thede2025wikibigedit}
in their original eight-interval order.
\method{} achieves the highest Original Match at 3K, 10K,
50K, and 100K among evaluated methods, with little fluctuation
(Table~\ref{tab:prefix-comparison}).
At 150K, it improves and surpasses Full-history
Replay (LoRA) on all four Match metrics, while the baseline's
Rephrase, Persona, and Locality scores decline
(Appendix~\ref{app:wikibigedit-150k}).
This suggests better generalization and less interference
as the stream grows.

\begin{figure*}[t]
\centering
\includegraphics[width=\textwidth]{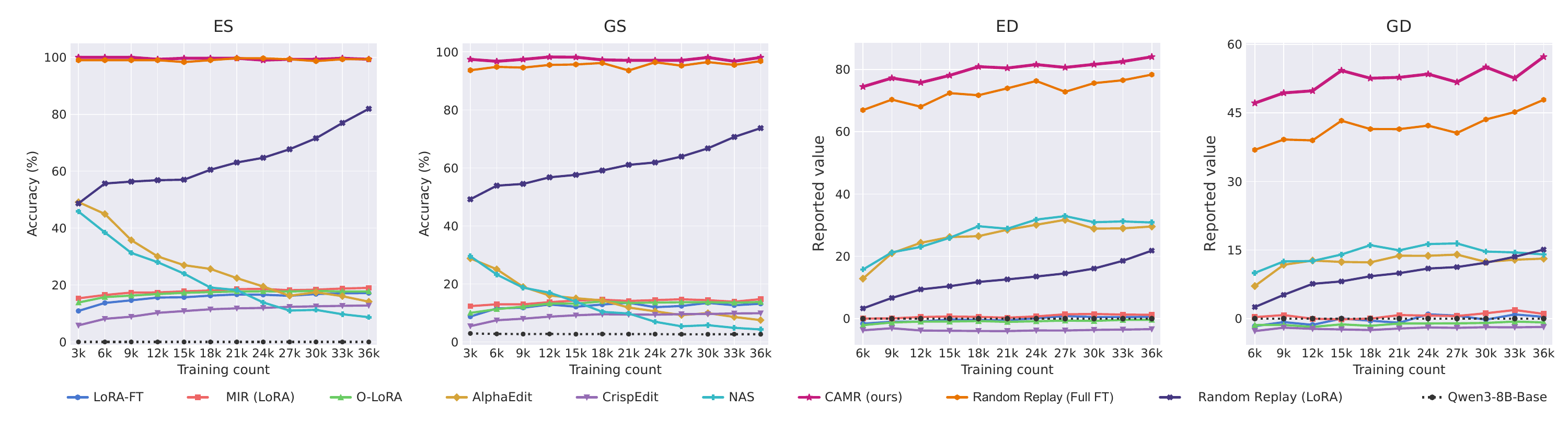}
\caption{WikiFactDiff replacement results over the 36K-update stream, evaluated on the current stage every 3K updates. ES/GS report success rates, while ED/GD report new-minus-obsolete answer-probability gaps.}
\label{fig:wikifactdiff}
\end{figure*}

On WikiFactDiff \citep{ammar-khodja-etal-2024-wikifactdiff-large},
we process twelve stages of 2,700 new facts and 300 replacements.
\method{} maintains current-stage ES near 100\% and
consistently exceeds Random Replay (Full FT) on GS
(Figure~\ref{fig:wikifactdiff}).
At 36K, both reach 99.33\% ES, while \method{} achieves
98.08\% GS, a 1.25-point improvement.
Positive mean ED and GD at every reported replacement
checkpoint indicate higher average probability for updated
answers than obsolete ones.

\begin{table*}[!t]
\centering
\scriptsize
\setlength{\tabcolsep}{2.0pt}
\renewcommand{\arraystretch}{1.12}
\definecolor{scoregain}{RGB}{34,139,105}
\definecolor{scoreloss}{RGB}{194,74,66}
\definecolor{scoreflat}{RGB}{120,120,120}
\definecolor{camrrow}{RGB}{235,243,253}
\newcommand{\scorechange}[2]{\,\scalebox{0.78}{\textcolor{#1}{(#2)}}}
\caption{
Continual learning performance (\%) and final checkpoint change on the four-task TRACE stream. T4 parentheses report changes from the stage when each task was first learned. The light peach shading
\protect\colorbox{bestscore}{\phantom{\rule{1.15ex}{1.15ex}}}
marks the highest value in each column, these values are also bold.
}
\label{tab:trace-four-task}
\colorlet{prefix3k}{traceblue}
\colorlet{prefix10k}{tracepink}
\colorlet{prefix50k}{tracepurple}
\colorlet{prefix100k}{tracegray}
\resizebox{\textwidth}{!}{%
\begin{tabular}{@{}l*{10}{c}@{}}
\toprule
\multirow{3}{*}{Method}
& \multicolumn{1}{>{\columncolor{prefix3k}}c}{T1}
& \multicolumn{2}{c}{\cellcolor{prefix10k}T2}
& \multicolumn{3}{c}{\cellcolor{prefix50k}T3}
& \multicolumn{4}{c}{\cellcolor{prefix100k}T4 (change from post-acquisition)} \\
& \multicolumn{1}{>{\columncolor{prefix3k}}c}{Acc.}
& \multicolumn{1}{>{\columncolor{prefix10k}}c}{Acc.}
& \multicolumn{1}{>{\columncolor{prefix10k}}c}{ROUGE-L}
& \multicolumn{1}{>{\columncolor{prefix50k}}c}{Acc.}
& \multicolumn{1}{>{\columncolor{prefix50k}}c}{ROUGE-L}
& \multicolumn{1}{>{\columncolor{prefix50k}}c}{Edit}
& \multicolumn{1}{>{\columncolor{prefix100k}}c}{Acc.}
& \multicolumn{1}{>{\columncolor{prefix100k}}c}{ROUGE-L}
& \multicolumn{1}{>{\columncolor{prefix100k}}c}{Edit}
& \multicolumn{1}{>{\columncolor{prefix100k}}c}{Acc.} \\
& \multicolumn{1}{>{\columncolor{prefix3k}}c}{C-STANCE}
& \multicolumn{1}{>{\columncolor{prefix10k}}c}{C-STANCE}
& \multicolumn{1}{>{\columncolor{prefix10k}}c}{MeetingBank}
& \multicolumn{1}{>{\columncolor{prefix50k}}c}{C-STANCE}
& \multicolumn{1}{>{\columncolor{prefix50k}}c}{MeetingBank}
& \multicolumn{1}{>{\columncolor{prefix50k}}c}{Py150}
& \multicolumn{1}{>{\columncolor{prefix100k}}c}{C-STANCE}
& \multicolumn{1}{>{\columncolor{prefix100k}}c}{MeetingBank}
& \multicolumn{1}{>{\columncolor{prefix100k}}c}{Py150}
& \multicolumn{1}{>{\columncolor{prefix100k}}c}{ScienceQA} \\
\midrule
\multicolumn{11}{c}{\textit{Fine-tuning and parameter-space methods}} \\
\midrule
Fine-Tuning (LoRA)
& 50.55 & 50.10 & 58.91 & 53.70 & 56.06 & 67.88 & 53.85\scorechange{scoregain}{+3.30} & 56.30\scorechange{scoreloss}{-2.61} & 67.30\scorechange{scoreloss}{-0.58} & 94.25 \\
Sequential Fine-Tuning (LoRA)
& 56.60 & 56.75 & 65.30 & 56.80 & 58.66 & 67.69 & 55.00\scorechange{scoreloss}{-1.60} & 53.85\scorechange{scoreloss}{-11.45} & 66.25\scorechange{scoreloss}{-1.44} & 91.95 \\
Sequential Fine-Tuning (Full FT)
& 49.30 & 46.05 & 69.20 & 46.75 & 68.14 & 63.39 & 47.70\scorechange{scoreloss}{-1.60} & 68.97\scorechange{scoreloss}{-0.23} & 62.76\scorechange{scoreloss}{-0.63} & 92.60 \\
Joint Fine-Tuning (Full FT)
& 46.25 & 51.65 & 42.70 & 53.60 & 44.74 & 60.34 & 51.50\scorechange{scoregain}{+5.25} & 68.16\scorechange{scoregain}{+25.46} & 59.60\scorechange{scoreloss}{-0.74} & 94.00 \\
O-LoRA
& 54.15 & 53.75 & 56.97 & 54.25 & 55.35 & 67.82 & 54.05\scorechange{scoreloss}{-0.10} & 56.40\scorechange{scoreloss}{-0.57} & 67.54\scorechange{scoreloss}{-0.28} & 93.30 \\
Adam-NSCL
& 50.30 & 50.25 & 17.01 & 50.30 & 17.06 & 11.48 & 50.30\scorechange{scoreflat}{+0.00} & 17.05\scorechange{scoregain}{+0.04} & 11.48\scorechange{scoreflat}{+0.00} & 87.65 \\
LoRA-Merge
& \best{58.20} & \best{58.75} & 32.65 & \best{59.00} & 33.53 & 20.83 & 56.60\scorechange{scoreloss}{-1.60} & 33.68\scorechange{scoregain}{+1.03} & 21.42\scorechange{scoregain}{+0.59} & 89.00 \\
\midrule
\multicolumn{11}{c}{\textit{Knowledge-editing methods}} \\
\midrule
LocFT
& 55.95 & 54.75 & 59.92 & 54.45 & 37.69 & 66.99 & 55.30\scorechange{scoreloss}{-0.65} & 31.95\scorechange{scoreloss}{-27.97} & 66.04\scorechange{scoreloss}{-0.95} & 88.95 \\
MEMIT
& 51.55 & 50.10 & 1.39 & 47.65 & 1.22 & 5.58 & 46.35\scorechange{scoreloss}{-5.20} & 1.21\scorechange{scoreloss}{-0.18} & 5.36\scorechange{scoreloss}{-0.22} & 88.85 \\
AlphaEdit
& 50.70 & 0.30 & 2.73 & 0.60 & 0.14 & 1.97 & 0.05\scorechange{scoreloss}{-50.65} & 0.09\scorechange{scoreloss}{-2.64} & 5.89\scorechange{scoregain}{+3.92} & 0.00 \\
NAS
& 50.35 & 44.40 & 17.03 & 40.70 & 15.73 & 5.11 & 33.75\scorechange{scoreloss}{-16.60} & 10.84\scorechange{scoreloss}{-6.19} & 5.85\scorechange{scoregain}{+0.74} & 0.75 \\
MEMIT + NAS
& 47.00 & 35.60 & 1.32 & 35.40 & 1.28 & 4.19 & 26.55\scorechange{scoreloss}{-20.45} & 1.18\scorechange{scoreloss}{-0.14} & 4.02\scorechange{scoreloss}{-0.17} & 64.10 \\
CrispEdit
& 56.35 & 43.90 & 33.27 & 43.15 & 31.60 & 62.62 & 47.40\scorechange{scoreloss}{-8.95} & 24.12\scorechange{scoreloss}{-9.15} & 62.01\scorechange{scoreloss}{-0.61} & 87.50 \\
UltraEdit
& 52.55 & 52.50 & 16.99 & 51.45 & 16.99 & 11.59 & 51.90\scorechange{scoreloss}{-0.65} & 17.08\scorechange{scoregain}{+0.09} & 11.68\scorechange{scoregain}{+0.09} & 87.30 \\
StableEdit
& 56.05 & 56.10 & 21.04 & 56.10 & 21.17 & 6.33 & 56.35\scorechange{scoregain}{+0.30} & 21.10\scorechange{scoregain}{+0.06} & 6.31\scorechange{scoreloss}{-0.02} & 87.65 \\
WISE-Retrieve
& 45.65 & 45.65 & 18.38 & 45.65 & 18.14 & 44.75 & 45.65\scorechange{scoreflat}{+0.00} & 18.13\scorechange{scoreloss}{-0.25} & 44.75\scorechange{scoreflat}{+0.00} & 85.30 \\
\midrule
\multicolumn{11}{c}{\textit{Replay-based continual-learning methods}} \\
\midrule
Random Replay (Full FT)
& 51.80 & 50.50 & 66.41 & 51.05 & 66.89 & 68.13 & 50.50\scorechange{scoreloss}{-1.30} & 66.45\scorechange{scoregain}{+0.04} & 67.80\scorechange{scoreloss}{-0.33} & 95.35 \\
Random Replay (LoRA)
& 54.80 & 55.60 & 63.38 & 56.05 & 60.93 & 67.55 & 54.45\scorechange{scoreloss}{-0.35} & 52.33\scorechange{scoreloss}{-11.05} & 65.52\scorechange{scoreloss}{-2.03} & 91.75 \\
Semantic Replay (LoRA)
& 53.45 & 55.05 & 64.96 & 54.60 & 53.99 & 67.13 & 51.90\scorechange{scoreloss}{-1.55} & 52.34\scorechange{scoreloss}{-12.62} & 66.67\scorechange{scoreloss}{-0.46} & 92.70 \\
TiC-style Replay (LoRA)
& 55.75 & 53.75 & 64.54 & 52.15 & 58.13 & 67.43 & 55.45\scorechange{scoreloss}{-0.30} & 51.04\scorechange{scoreloss}{-13.50} & 67.12\scorechange{scoreloss}{-0.31} & 91.30 \\
Maximally Interfered Retrieval (LoRA)
& 53.75 & 52.90 & 63.29 & 52.05 & 49.21 & \best{68.39} & 50.30\scorechange{scoreloss}{-3.45} & 35.10\scorechange{scoreloss}{-28.19} & 66.32\scorechange{scoreloss}{-2.07} & 91.05 \\
Transfer Selective Replay (LoRA)
& 56.60 & 55.85 & 62.83 & 55.95 & 23.75 & 62.22 & 55.35\scorechange{scoreloss}{-1.25} & 41.40\scorechange{scoreloss}{-21.43} & 58.21\scorechange{scoreloss}{-4.01} & 92.55 \\
\rowcolor{camrrow}
\textbf{CAMR (Full FT, ours)}
& 57.50 & 55.65 & \best{70.39} & 56.75 & \best{69.22} & 68.34 & \best{56.75}\scorechange{scoreloss}{-0.75} & \best{69.84}\scorechange{scoreloss}{-0.55} & \best{67.92}\scorechange{scoreloss}{-0.42} & \best{95.40} \\
\rowcolor{camrrow}
\textbf{CAMR (LoRA, ours)}
& 54.60 & 56.70 & 64.30
& 54.75 & 65.38 & 66.49
& 55.00\scorechange{scoregain}{+0.40} & 63.10\scorechange{scoreloss}{-1.20} & 64.45\scorechange{scoreloss}{-2.04} & 92.80 \\
\bottomrule
\end{tabular}%
}
\end{table*}

\subsection{Cross-Task and Cross-Domain Performance}

On the TRACE sequence C-STANCE $\rightarrow$ MeetingBank
$\rightarrow$ Py150 $\rightarrow$ text-only ScienceQA,
\method{} achieves the highest final scores among evaluated
methods on all four tasks: 56.75\%, 69.84\%, 67.92\%,
and 95.40\%, respectively (Table~\ref{tab:trace-four-task}).
The three earlier tasks lose only 0.75, 0.55, and 0.42 points
relative to their post-acquisition scores, indicating
limited forgetting across successive tasks.

\begin{table*}[!t]
\centering
\scriptsize
\setlength{\tabcolsep}{2.0pt}
\renewcommand{\arraystretch}{1.12}
\definecolor{scoregain}{RGB}{34,139,105}
\definecolor{scoreloss}{RGB}{194,74,66}
\definecolor{scoreflat}{RGB}{120,120,120}
\definecolor{camrrow}{RGB}{235,243,253}
\newcommand{\scorechange}[2]{\,\scalebox{0.78}{\textcolor{#1}{(#2)}}}
\caption{
Continual learning performance (\%) and final checkpoint change on the four-task cross-domain stream. T4 parentheses report changes from the stage when each task was first learned. The light peach shading
\protect\colorbox{bestscore}{\phantom{\rule{1.15ex}{1.15ex}}}
marks the highest value in each column, these values are also bold.
}
\label{tab:cross-domain-continual-results}

\resizebox{\textwidth}{!}{%
\begin{tabular}{
  @{}l*{10}{c}@{}
}
\toprule
\multirow{3}{*}{\textbf{Method}}
& \multicolumn{1}{c}{\cellcolor{prefix3k}\textbf{T1}}
& \multicolumn{2}{c}{\cellcolor{prefix10k}\textbf{T2}}
& \multicolumn{3}{c}{\cellcolor{prefix50k}\textbf{T3}}
& \multicolumn{4}{c}{\cellcolor{prefix100k}\textbf{T4 (change from post-acquisition)}} \\

& \multicolumn{1}{>{\columncolor{prefix3k}}c}{Acc.}
& \multicolumn{1}{>{\columncolor{prefix10k}}c}{Acc.}
& \multicolumn{1}{>{\columncolor{prefix10k}}c}{Acc.}
& \multicolumn{1}{>{\columncolor{prefix50k}}c}{Acc.}
& \multicolumn{1}{>{\columncolor{prefix50k}}c}{Acc.}
& \multicolumn{1}{>{\columncolor{prefix50k}}c}{pass@1}
& \multicolumn{1}{>{\columncolor{prefix100k}}c}{Acc.}
& \multicolumn{1}{>{\columncolor{prefix100k}}c}{Acc.}
& \multicolumn{1}{>{\columncolor{prefix100k}}c}{pass@1}
& \multicolumn{1}{>{\columncolor{prefix100k}}c}{ROUGE-L} \\

& \multicolumn{1}{>{\columncolor{prefix3k}}c}{FinQA}
& \multicolumn{1}{>{\columncolor{prefix10k}}c}{FinQA}
& \multicolumn{1}{>{\columncolor{prefix10k}}c}{GSM8K}
& \multicolumn{1}{>{\columncolor{prefix50k}}c}{FinQA}
& \multicolumn{1}{>{\columncolor{prefix50k}}c}{GSM8K}
& \multicolumn{1}{>{\columncolor{prefix50k}}c}{APPS}
& \multicolumn{1}{>{\columncolor{prefix100k}}c}{FinQA}
& \multicolumn{1}{>{\columncolor{prefix100k}}c}{GSM8K}
& \multicolumn{1}{>{\columncolor{prefix100k}}c}{APPS}
& \multicolumn{1}{>{\columncolor{prefix100k}}c}{SAMSum} \\

\midrule
\multicolumn{11}{c}{
  \textit{Fine-tuning and parameter-space methods}
} \\
\midrule

Fine-Tuning (LoRA)
& 30.51 & 22.23 & 82.41 & 26.94 & 80.89 & 30.40
& 22.23\scorechange{scoreloss}{-8.28} & 70.05\scorechange{scoreloss}{-12.36} & 30.60\scorechange{scoregain}{+0.20} & 43.79 \\

Sequential Fine-Tuning (Full FT)
& 23.37 & 24.24 & 75.06 & 24.85 & 73.62 & 20.80
& 23.54\scorechange{scoregain}{+0.17} & 70.66\scorechange{scoreloss}{-4.40} & 22.80\scorechange{scoregain}{+2.00} & 42.25 \\

O-LoRA
& 31.56 & 31.39 & 82.94 & 30.95 & 82.64 & 28.80
& 30.43\scorechange{scoreloss}{-1.13} & 83.55\scorechange{scoregain}{+0.61} & 29.10\scorechange{scoregain}{+0.30} & 44.59 \\

Adam-NSCL
& 7.15 & 7.15 & 75.66
& 7.15 & 75.44 & \best{36.10}
& 7.06\scorechange{scoreloss}{-0.09} & 76.04\scorechange{scoregain}{+0.38} & \best{36.10}\scorechange{scoreflat}{+0.00} & 25.46 \\

LoRA-Merge
& 64.25 & 64.17 & 82.49 & 64.17 & 83.70 & 33.50
& 63.30\scorechange{scoreloss}{-0.95} & 83.32\scorechange{scoregain}{+0.83} & 32.60\scorechange{scoreloss}{-0.90} & 41.17 \\

\midrule
\multicolumn{11}{c}{
  \textit{Knowledge-editing methods}
} \\
\midrule

LocFT
& 74.02 & 67.57 & \best{83.62} & 67.57 & 79.61 & 32.20
& 68.09\scorechange{scoreloss}{-5.93} & 81.27\scorechange{scoreloss}{-2.35} & 32.80\scorechange{scoregain}{+0.60} & 43.82 \\

MEMIT
& 4.80
& 6.63 & 80.67
& 6.71 & 80.59 & 23.80
& 6.19\scorechange{scoregain}{+1.39} & 80.74\scorechange{scoregain}{+0.07} & 24.70\scorechange{scoregain}{+0.90} & 32.47 \\

AlphaEdit
& 6.54
& 6.89 & 67.02
& 6.97 & 66.94 & 32.60
& 6.97\scorechange{scoregain}{+0.43} & 67.85\scorechange{scoregain}{+0.83} & 32.60\scorechange{scoreflat}{+0.00} & 26.19 \\

NAS
& 7.50 & 7.32 & 74.37 & 7.59 & 74.07 & 35.20
& 7.41\scorechange{scoreloss}{-0.09} & 75.36\scorechange{scoregain}{+0.99} & 35.60\scorechange{scoregain}{+0.40} & 25.72 \\

MEMIT + NAS
& 7.32 & 7.59 & 78.85 & 7.41 & 78.85 & 34.80
& 7.67\scorechange{scoregain}{+0.35} & 78.77\scorechange{scoreloss}{-0.08} & 35.90\scorechange{scoregain}{+1.10} & 25.68 \\

CrispEdit
& 50.22 & 16.74 & 81.88 & 20.92 & 78.24 & 25.60
& 21.10\scorechange{scoreloss}{-29.12} & 48.37\scorechange{scoreloss}{-33.51} & 7.50\scorechange{scoreloss}{-18.10} & 37.96 \\

UltraEdit
& 8.02
& 8.20 & 80.52
& 8.02 & 78.62 & 34.50
& 8.11\scorechange{scoregain}{+0.09} & 80.21\scorechange{scoreloss}{-0.31} & 34.10\scorechange{scoreloss}{-0.40} & 26.27 \\

StableEdit
& 17.87 & 17.52 & 83.55 & 17.87 & 82.87 & 16.70
& 17.52\scorechange{scoreloss}{-0.35} & 82.56\scorechange{scoreloss}{-0.99} & 14.90\scorechange{scoreloss}{-1.80} & 42.58 \\

WISE-Retrieve
& 2.09 & 2.09 & 47.38
& 2.18 & 59.89 & 1.70
& 1.92\scorechange{scoreloss}{-0.17} & 58.98\scorechange{scoregain}{+11.60} & 2.00\scorechange{scoregain}{+0.30} & 12.96 \\

\midrule
\multicolumn{11}{c}{
  \textit{Replay-based continual-learning methods}
} \\
\midrule

Random Replay (Full FT)
& 75.15 & 74.80 & 81.88 & 73.67 & 78.62 & 26.30
& 73.41\scorechange{scoreloss}{-1.74} & 76.12\scorechange{scoreloss}{-5.76} & 21.40\scorechange{scoreloss}{-4.90} & 41.67 \\

Random Replay (LoRA)
& 32.17 & 35.14 & 82.34 & 35.22 & 80.74 & 30.70
& 26.85\scorechange{scoreloss}{-5.32} & 76.65\scorechange{scoreloss}{-5.69} & 27.60\scorechange{scoreloss}{-3.10} & 44.86 \\

Semantic Replay (LoRA)
& 28.86 & 34.70 & 80.06 & 34.26 & 80.06 & 29.60
& 30.95\scorechange{scoregain}{+2.09} & 77.94\scorechange{scoreloss}{-2.12} & 29.20\scorechange{scoreloss}{-0.40} & 43.75 \\

Relation Oracle Replay (Full FT)
& 27.72 & 33.83 & 82.56 & 33.83 & 82.18 & 29.30
& 29.64\scorechange{scoregain}{+1.92} & 78.77\scorechange{scoreloss}{-3.79} & 27.70\scorechange{scoreloss}{-1.60} & 43.43 \\

TiC-style Replay (LoRA)
& 36.79 & 35.14 & 83.02 & 34.87 & 81.43 & 28.90
& 34.18\scorechange{scoreloss}{-2.61} & 81.20\scorechange{scoreloss}{-1.82} & 30.80\scorechange{scoregain}{+1.90} & 44.94 \\

Maximally Interfered Retrieval (LoRA)
& 32.61 & 32.61 & 82.26 & 31.12 & 81.73 & 27.90
& 31.30\scorechange{scoreloss}{-1.31} & 82.26\scorechange{scoreflat}{+0.00} & 29.70\scorechange{scoregain}{+1.80} & 43.61 \\

Transfer Selective Replay (LoRA)
& 36.97 & 36.09 & 83.24 & 35.22 & 82.03 & 30.80
& 34.52\scorechange{scoreloss}{-2.45} & 81.43\scorechange{scoreloss}{-1.81} & 29.60\scorechange{scoreloss}{-1.20} & 44.26 \\

\rowcolor{camrrow}
\textbf{CAMR (Full FT, ours)}
& 75.24 & 75.85 & 81.58
& 75.94 & 82.18 & 30.10
& 76.11\scorechange{scoregain}{+0.87} & 81.58\scorechange{scoreflat}{+0.00} & 30.50\scorechange{scoregain}{+0.40} & \best{45.10} \\

\rowcolor{camrrow}
\textbf{CAMR (LoRA, ours)}
& \best{76.10} & \best{76.36} & 83.47
& \best{77.15} & \best{85.75} & 32.80
& \best{77.24}\scorechange{scoregain}{+1.14} & \best{84.91}\scorechange{scoregain}{+1.44} & 31.80\scorechange{scoreloss}{-1.00} & 43.96 \\

\bottomrule
\end{tabular}%
}
\end{table*}

On FinQA $\rightarrow$ GSM8K $\rightarrow$ APPS introductory
$\rightarrow$ SAMSum, all three earlier tasks retain or
improve upon their post-acquisition scores
(Table~\ref{tab:cross-domain-continual-results}).
Final scores reach 76.11\%, 81.58\%, 30.50\%, and 45.10\%,
exceeding Random Replay (Full FT) by 2.70, 5.46, 9.10,
and 3.43 points, respectively.
\method{} leads on FinQA and SAMSum, while O-LoRA
and Adam-NSCL achieve the highest GSM8K and APPS scores,
respectively.

\subsection{Sensitivity to Replay Count}
\label{sec:budget_comparison}

\begin{table}[!htb]
\centering
\caption{Replay strategies under different per-arrival replay caps $K$. All methods use full-parameter fine-tuning. Best scores within each $K$ group are bold.}
\label{tab:k-ablation}

\renewcommand{\arraystretch}{1.15}
\setlength{\tabcolsep}{4pt}

\resizebox{\columnwidth}{!}{%
\begin{tabular}{clcccc}
\toprule
$K$
& \textbf{Method}
& \textbf{Original}
& \textbf{Rephrase}
& \textbf{Persona}
& \textbf{Locality}
\\
\midrule

\multirow{3}{*}{4}
& \method{} (ours)
& \textbf{93.98\%}
& \textbf{86.56\%}
& 69.98\%
& 18.34\%
\\
& Random Replay
& 93.62\%
& 86.08\%
& \textbf{70.74\%}
& \textbf{19.58\%}
\\
& Semantic Replay
& 80.40\%
& 73.36\%
& 63.21\%
& 11.91\%
\\

\midrule

\multirow{3}{*}{8}
& \method{} (ours)
& \textbf{96.70\%}
& \textbf{89.26\%}
& \textbf{72.62\%}
& \textbf{19.86\%}
\\
& Random Replay
& 96.62\%
& 89.08\%
& 70.92\%
& 19.30\%
\\
& Semantic Replay
& 82.62\%
& 74.73\%
& 65.70\%
& 13.67\%
\\

\midrule

\multirow{3}{*}{16}
& \method{} (ours)
& \textbf{98.10\%}
& \textbf{90.42\%}
& 71.72\%
& \textbf{19.16\%}
\\
& Random Replay
& 97.52\%
& 88.16\%
& 68.74\%
& 18.98\%
\\
& Semantic Replay
& 87.96\%
& 79.42\%
& \textbf{71.78\%}
& 14.06\%
\\

\bottomrule
\end{tabular}%
}
\end{table}

Table~\ref{tab:k-ablation} evaluates different replay counts.
\method{} leads Original and Rephrase Match at every tested
budget, showing that its advantage is not tied to a single $K$.
Increasing $K$ from 4 to 16 improves these scores from
93.98\%/86.56\% to 98.10\%/90.42\%.
At $K=8$, \method{} achieves its highest Persona and Locality
scores (72.62\%/19.86\%) and leads all four metrics.
We use $K=4$ as the default to evaluate the method
under a limited replay budget.

\subsection{Writer Component Ablations}
\label{sec:ablation}
Table~\ref{tab:strategy-ablation} compares the complete
writer with five ablations.
Specifically, w/o Action Selection uses \addact{}-only routing;
w/o Online Feedback freezes the initialized writer;
w/o Memory Gate weights all candidates equally;
w/o \ignoreact{} masks that action;
and w/o Replay on \ignoreact{} skips historical replay for that action.
The complete writer achieves the best performance on
all four tasks, tying w/o Online Feedback on NumGLUE-cm.
Removing action selection or \ignoreact{} degrades
performance across all tasks, supporting the benefit
of adapting current-example supervision and historical replay.
NumGLUE-ds is particularly sensitive to memory gating
and replay under \ignoreact{}, with substantial degradation
when either component is removed.
Online feedback improves three tasks, with the largest
gain on NumGLUE-ds.

\begin{table}[!htb]
\centering
\caption{
Writer ablation results under full-parameter fine-tuning.
The light peach shading
\protect\colorbox{bestscore}{\phantom{\rule{1.15ex}{1.15ex}}}
marks the highest score in each column, including ties.
}
\label{tab:strategy-ablation}

\resizebox{\columnwidth}{!}{%
\begin{tabular}{lcccc}
\toprule

\textbf{Variant}
& \textbf{FOMC}
& \textbf{NumGLUE-cm}
& \textbf{NumGLUE-ds}
& \textbf{Py150}
\\

\midrule

\method{} (ours)
& \best{72.38}
& \best{76.54}
& \best{77.54}
& \best{67.75}
\\

w/o Action Selection
& 71.25
& 74.07
& 74.57
& 66.89
\\

w/o Online Feedback
& 71.25
& \best{76.54}
& 75.69
& 67.58
\\

w/o Memory Gate
& 70.63
& 74.07
& 72.85
& 66.65
\\

w/o \ignoreact{}
& 71.88
& 74.07
& 76.63
& 66.63
\\

w/o Replay on \ignoreact{}
& 71.25
& 72.84
& 73.54
& 66.49
\\

\bottomrule
\end{tabular}%
}

\end{table}

\subsection{Replay Sample Selection Analysis}
\label{sec:sample_selection}
\method{} balances focused replay with historical coverage
under a fixed budget.
At 100K, its replay share for the high-frequency core-3K
set is 2.42$\times$ that of Random Replay
(Table~\ref{tab:wikibigedit-core3k-replay-share}).
Its replay concentrates on regions similar to Semantic
Replay but covers more arrived facts: 90.4\% versus 83.9\%
(Appendix~\ref{app:semantic-trajectory}).
This pattern is consistent with its hybrid sampling:
\mergeact{} prioritizes compatible related history while
reserving uniform slots, and \addact{} samples uniformly.
\method{} also outperforms Semantic Replay on Original
and Rephrase Match at every tested $K$
(Table~\ref{tab:k-ablation}).

\section{Conclusion}
We introduced \method{}, which jointly controls
current-example supervision and bounded historical replay
through compatibility-aware memory and an online writer policy.
Experiments on factual updating, cross-task learning,
and cross-domain learning demonstrate strong acquisition
and retention across replay budgets.
Our results highlight the value of jointly adapting
current supervision and historical replay in continual learning.



 \section*{Limitations}
  Due to computational constraints, our experiments focus on Qwen3-8B-Base; since \method{} is not tied to a specific architecture or scale, we plan to extend it to larger models and other model families. Some baselines use LoRA rather than
  full-parameter fine-tuning to limit cost, and a full-parameter comparison of all baselines is left for future work. \ignoreact{} is not triggered on WikiBigEdit, which contains few repeated facts, but it accounted for 19.14\% of decisions under the gate defined in Section~\ref{sec:policy} on the
  APPS stream with repeated examples (Appendix~\ref{app:writer-action-utilization}). Removing \ignoreact{} lowers performance on all four ablation tasks, showing its benefit on redundant data. We will further evaluate it on streams with naturally occurring repetitions.

\FloatBarrier
\bibliography{reference}

@article{cheng2026conditional,
  title = {Conditional Memory via Scalable Lookup: A New Axis of Sparsity for Large Language Models},
  author = {Cheng, Xin and Tian, Rui and Zeng, Wangding and Dai, Damai and Chen, Qinyu and Wang, Bingxuan and Xie, Zhenda and Huang, Kezhao and Yu, Xingkai and Deng, Chengqi and Zhou, Shangyan and Zhao, Chenggang and Hao, Zhewen and Li, Yukun and Zhang, Han and Zhang, Zhengyan and Wei, Yixuan and Xu, M.Y. and Zhang, Huishuai and Zhao, Dongyan and Liang, Wenfeng},
  journal = {arXiv preprint arXiv:2601.07372},
  year = {2026},
  doi = {10.48550/arXiv.2601.07372},
  eprint = {2601.07372},
  archiveprefix = {arXiv},
  primaryclass = {cs.CL},
  url = {https://arxiv.org/abs/2601.07372}
}

@inproceedings{song2020mpnet,
  title = {{MPNet}: Masked and Permuted Pre-training for Language Understanding},
  author = {Song, Kaitao and Tan, Xu and Qin, Tao and Lu, Jianfeng and Liu, Tie-Yan},
  booktitle = {Advances in Neural Information Processing Systems},
  volume = {33},
  pages = {16857--16867},
  year = {2020},
  publisher = {Curran Associates, Inc.},
  url = {https://proceedings.neurips.cc/paper_files/paper/2020/hash/c3a690be93aa602ee2dc0ccab5b7b67e-Abstract.html}
}

@inproceedings{mnih2016asynchronous,
  title = {Asynchronous Methods for Deep Reinforcement Learning},
  author = {Mnih, Volodymyr and Badia, Adria Puigdomenech and Mirza, Mehdi and Graves, Alex and Lillicrap, Timothy and Harley, Tim and Silver, David and Kavukcuoglu, Koray},
  booktitle = {Proceedings of The 33rd International Conference on Machine Learning},
  series = {Proceedings of Machine Learning Research},
  volume = {48},
  pages = {1928--1937},
  year = {2016},
  publisher = {PMLR},
  url = {https://proceedings.mlr.press/v48/mniha16.html}
}

@incollection{mccloskey1989catastrophic,
  title = {Catastrophic Interference in Connectionist Networks: The Sequential Learning Problem},
  author = {McCloskey, Michael and Cohen, Neal J.},
  booktitle = {Psychology of Learning and Motivation},
  volume = {24},
  pages = {109--165},
  publisher = {Academic Press},
  year = {1989},
  doi = {10.1016/S0079-7421(08)60536-8},
  url = {https://doi.org/10.1016/S0079-7421(08)60536-8}
}

@article{delange2022survey,
  title = {A Continual Learning Survey: Defying Forgetting in Classification Tasks},
  author = {De Lange, Matthias and Aljundi, Rahaf and Masana, Marc and Parisot, Sarah and Jia, Xu and Leonardis, Ales and Slabaugh, Gregory and Tuytelaars, Tinne},
  journal = {IEEE Transactions on Pattern Analysis and Machine Intelligence},
  volume = {44},
  number = {7},
  pages = {3366--3385},
  year = {2022},
  doi = {10.1109/TPAMI.2021.3057446},
  url = {https://doi.org/10.1109/TPAMI.2021.3057446}
}

@inproceedings{rolnick2019experience,
  title = {Experience Replay for Continual Learning},
  author = {Rolnick, David and Ahuja, Arun and Schwarz, Jonathan and Lillicrap, Timothy and Wayne, Greg},
  booktitle = {Advances in Neural Information Processing Systems},
  volume = {32},
  year = {2019},
  publisher = {Curran Associates, Inc.},
  url = {https://proceedings.neurips.cc/paper_files/paper/2019/hash/fa7cdfad1a5aaf8370ebeda47a1ff1c3-Abstract.html}
}

@inproceedings{aljundi2019online,
  title = {Online Continual Learning with Maximal Interfered Retrieval},
  author = {Aljundi, Rahaf and Belilovsky, Eugene and Tuytelaars, Tinne and Charlin, Laurent and Caccia, Massimo and Lin, Min and Page-Caccia, Lucas},
  booktitle = {Advances in Neural Information Processing Systems},
  volume = {32},
  year = {2019},
  publisher = {Curran Associates, Inc.},
  url = {https://proceedings.neurips.cc/paper_files/paper/2019/hash/15825aee15eb335cc13f9b559f166ee8-Abstract.html}
}

@inproceedings{meng2022locating,
  title = {Locating and Editing Factual Associations in {GPT}},
  author = {Meng, Kevin and Bau, David and Andonian, Alex and Belinkov, Yonatan},
  booktitle = {Advances in Neural Information Processing Systems},
  volume = {35},
  pages = {17359--17372},
  year = {2022},
  publisher = {Curran Associates, Inc.},
  doi = {10.52202/068431-1262},
  url = {https://proceedings.neurips.cc/paper_files/paper/2022/hash/6f1d43d5a82a37e89b0665b33bf3a182-Abstract-Conference.html}
}

@inproceedings{meng2023mass,
  title = {Mass-Editing Memory in a Transformer},
  author = {Meng, Kevin and Sharma, Arnab Sen and Andonian, Alex and Belinkov, Yonatan and Bau, David},
  booktitle = {International Conference on Learning Representations},
  year = {2023},
  url = {https://openreview.net/forum?id=MkbcKIqIkl2},
  eprint = {2210.07229},
  archiveprefix = {arXiv}
}

@inproceedings{wise2024,
  title = {{WISE}: Rethinking the Knowledge Memory for Lifelong Model Editing of Large Language Models},
  author = {Wang, Peng and Li, Zexi and Zhang, Ningyu and Xu, Ziwen and Yao, Yunzhi and Jiang, Yong and Xie, Pengjun and Huang, Fei and Chen, Huajun},
  booktitle = {Advances in Neural Information Processing Systems},
  volume = {37},
  pages = {53764--53797},
  year = {2024},
  publisher = {Curran Associates, Inc.},
  doi = {10.52202/079017-1703},
  url = {https://proceedings.neurips.cc/paper_files/paper/2024/hash/60960ad78868fce5c165295fbd895060-Abstract-Conference.html}
}

@inproceedings{alphaedit2025,
  title = {{AlphaEdit}: Null-Space Constrained Knowledge Editing for Language Models},
  author = {Fang, Junfeng and Jiang, Houcheng and Wang, Kun and Ma, Yunshan and Jie, Shi and Wang, Xiang and He, Xiangnan and Chua, Tat-Seng},
  booktitle = {International Conference on Learning Representations},
  year = {2025},
  note = {Oral presentation},
  eprint = {2410.02355},
  archiveprefix = {arXiv},
  doi = {10.48550/arXiv.2410.02355},
  url = {https://arxiv.org/abs/2410.02355}
}

@inproceedings{gupta2024rrome,
  title = {Rebuilding {ROME}: Resolving Model Collapse during Sequential Model Editing},
  author = {Gupta, Akshat and Baskaran, Sidharth and Anumanchipalli, Gopala},
  booktitle = {Proceedings of the 2024 Conference on Empirical Methods in Natural Language Processing},
  pages = {21738--21744},
  year = {2024},
  publisher = {Association for Computational Linguistics},
  doi = {10.18653/v1/2024.emnlp-main.1210},
  url = {https://aclanthology.org/2024.emnlp-main.1210/}
}

@article{kirkpatrick2017ewc,
  title = {Overcoming Catastrophic Forgetting in Neural Networks},
  author = {Kirkpatrick, James and Pascanu, Razvan and Rabinowitz, Neil and Veness, Joel and Desjardins, Guillaume and Rusu, Andrei A. and Milan, Kieran and Quan, John and Ramalho, Tiago and Grabska-Barwinska, Agnieszka and Hassabis, Demis and Clopath, Claudia and Kumaran, Dharshan and Hadsell, Raia},
  journal = {Proceedings of the National Academy of Sciences},
  volume = {114},
  number = {13},
  pages = {3521--3526},
  year = {2017},
  doi = {10.1073/pnas.1611835114},
  url = {https://doi.org/10.1073/pnas.1611835114}
}

@inproceedings{charikar2002similarity,
  title = {Similarity Estimation Techniques from Rounding Algorithms},
  author = {Charikar, Moses S.},
  booktitle = {Proceedings of the 34th Annual ACM Symposium on Theory of Computing},
  pages = {380--388},
  year = {2002},
  month = may,
  publisher = {ACM},
  doi = {10.1145/509907.509965},
  url = {https://doi.org/10.1145/509907.509965}
}

@inproceedings{zhang2019rmsnorm,
  title = {Root Mean Square Layer Normalization},
  author = {Zhang, Biao and Sennrich, Rico},
  booktitle = {Advances in Neural Information Processing Systems},
  volume = {32},
  year = {2019},
  publisher = {Curran Associates, Inc.},
  url = {https://proceedings.neurips.cc/paper_files/paper/2019/hash/1e8a19426224ca89e83cef47f1e7f53b-Abstract.html}
}

@inproceedings{loshchilov2019adamw,
  title = {Decoupled Weight Decay Regularization},
  author = {Loshchilov, Ilya and Hutter, Frank},
  booktitle = {International Conference on Learning Representations},
  year = {2019},
  url = {https://openreview.net/forum?id=Bkg6RiCqY7},
  note = {arXiv:1711.05101}
}

@article{trace2023,
  title={{TRACE}: A Comprehensive Benchmark for Continual Learning in Large Language Models},
  author={Wang, Xiao and Zhang, Yuansen and Chen, Tianze and Gao, Songyang and Jin, Senjie and Yang, Xianjun and Xi, Zhiheng and Zheng, Rui and Zou, Yicheng and Gui, Tao and Zhang, Qi and Huang, Xuanjing},
  journal={arXiv preprint arXiv:2310.06762}, year={2023},
  url={https://arxiv.org/abs/2310.06762}, eprint={2310.06762}, archivePrefix={arXiv}, primaryClass={cs.CL}
}

@inproceedings{cstance2023,
 title={{C}-{STANCE}: A Large Dataset for {C}hinese Zero-Shot Stance Detection},
 author={Zhao, Chenye and Li, Yingjie and Caragea, Cornelia},
 booktitle={Proceedings of the 61st Annual Meeting of the Association for Computational Linguistics (Volume 1: Long Papers)}, pages={13369--13385}, year={2023}, publisher={ACL}, doi={10.18653/v1/2023.acl-long.747}, url={https://aclanthology.org/2023.acl-long.747/}}

@inproceedings{meetingbank2023,
 title={{MeetingBank}: A Benchmark Dataset for Meeting Summarization},
 author={Hu, Yebowen and Ganter, Timothy and Deilamsalehy, Hanieh and Dernoncourt, Franck and Foroosh, Hassan and Liu, Fei},
 booktitle={Proceedings of the 61st Annual Meeting of the Association for Computational Linguistics (Volume 1: Long Papers)}, pages={16409--16423}, year={2023}, publisher={ACL}, doi={10.18653/v1/2023.acl-long.906}, url={https://aclanthology.org/2023.acl-long.906/}}

@inproceedings{py1502016,
 title={Probabilistic Model for Code with Decision Trees}, author={Raychev, Veselin and Bielik, Pavol and Vechev, Martin},
 booktitle={Proceedings of the 2016 ACM SIGPLAN International Conference on Object-Oriented Programming, Systems, Languages, and Applications}, pages={731--747}, year={2016}, publisher={ACM}, doi={10.1145/2983990.2984041}}

@inproceedings{scienceqa2022,
 title={Learn to Explain: Multimodal Reasoning via Thought Chains for Science Question Answering},
 author={Lu, Pan and Mishra, Swaroop and Xia, Tanglin and Qiu, Liang and Chang, Kai-Wei and Zhu, Song-Chun and Tafjord, Oyvind and Clark, Peter and Kalyan, Ashwin},
 booktitle={Advances in Neural Information Processing Systems}, volume={35}, pages={2507--2521}, year={2022}, url={https://doi.org/10.52202/068431-0182}}

@inproceedings{finqa2021,
 title={FinQA: A Dataset of Numerical Reasoning over Financial Data}, author={Chen, Zhiyu and Chen, Wenhu and Smiley, Charese and Shah, Sameena and Borova, Iana and Langdon, Dylan and Moussa, Reema and Beane, Matt and Huang, Ting-Hao and Routledge, Bryan and Wang, William Yang},
 booktitle={Proceedings of the 2021 Conference on Empirical Methods in Natural Language Processing}, pages={3697--3711}, year={2021}, publisher={ACL}, doi={10.18653/v1/2021.emnlp-main.300}, url={https://aclanthology.org/2021.emnlp-main.300/}}

@inproceedings{gsm8k2021,
 title={Training Verifiers to Solve Math Word Problems}, author={Cobbe, Karl and Kosaraju, Vineet and Bavarian, Mohammad and Chen, Mark and Jun, Heewoo and Kaiser, Lukasz and Plappert, Matthias and Tworek, Jerry and Hilton, Jacob and Nakano, Reiichiro and Hesse, Christopher and Schulman, John},
 booktitle={arXiv preprint arXiv:2110.14168}, year={2021}, eprint={2110.14168}, archivePrefix={arXiv}, url={https://arxiv.org/abs/2110.14168}}

@inproceedings{apps2021,
 title={Measuring Coding Challenge Competence With APPS}, author={Hendrycks, Dan and Basart, Steven and Kadavath, Saurav and Mazeika, Mantas and Arora, Akul and Guo, Ethan and Burns, Collin and Puranik, Samir and He, Horace and Song, Dawn and Steinhardt, Jacob},
 booktitle={arXiv preprint arXiv:2105.09938}, year={2021}, eprint={2105.09938}, archivePrefix={arXiv}, url={https://arxiv.org/abs/2105.09938}}

@inproceedings{samsum2019,
 title={{SAMSum} Corpus: A Human-annotated Dialogue Dataset for Abstractive Summarization}, author={Gliwa, Bogdan and Mochol, Iwona and Biesek, Maciej and Wawer, Aleksander},
 booktitle={Proceedings of the 2nd Workshop on New Frontiers in Summarization}, pages={70--79}, year={2019}, publisher={ACL}, doi={10.18653/v1/D19-5409}, url={https://aclanthology.org/D19-5409/}}

@inproceedings{fomc2023,
 title={Trillion Dollar Words: A New Financial Dataset, Task and Market Analysis}, author={Shah, Agam and Paturi, Suvan and Chava, Sudheer},
 booktitle={Proceedings of the 61st Annual Meeting of the Association for Computational Linguistics (Volume 1: Long Papers)}, pages={6664--6679}, year={2023}, publisher={ACL}, doi={10.18653/v1/2023.acl-long.368}, url={https://aclanthology.org/2023.acl-long.368/}}

@inproceedings{numglue2022,
 title={{NumGLUE}: A Suite of Fundamental yet Challenging Mathematical Reasoning Tasks}, author={Mishra, Swaroop and Mitra, Arindam and Varshney, Neeraj and Sachdeva, Bhavdeep and Clark, Peter and Baral, Chitta and Kalyan, Ashwin},
 booktitle={Proceedings of the 60th Annual Meeting of the Association for Computational Linguistics (Volume 1: Long Papers)}, pages={3505--3523}, year={2022}, publisher={ACL}, doi={10.18653/v1/2022.acl-long.246}, url={https://aclanthology.org/2022.acl-long.246/}}

@inproceedings{lora2022,
 title={{LoRA}: Low-Rank Adaptation of Large Language Models}, author={Hu, Edward J. and Shen, Yelong and Wallis, Phillip and Allen-Zhu, Zeyuan and Li, Yuanzhi and Wang, Shean and Wang, Lu and Chen, Weizhu}, booktitle={International Conference on Learning Representations}, year={2022}, url={https://openreview.net/forum?id=nZeVKeeFYf9}, eprint={2106.09685}}

@inproceedings{olora2023,
 title={Orthogonal Subspace Learning for Language Model Continual Learning}, author={Wang, Xiao and Chen, Tianze and Ge, Qiming and Xia, Han and Bao, Rong and Zheng, Rui and Zhang, Qi and Gui, Tao and Huang, Xuanjing}, booktitle={Findings of the Association for Computational Linguistics: EMNLP 2023}, pages={10658--10671}, year={2023}, publisher={ACL}, doi={10.18653/v1/2023.findings-emnlp.715}, url={https://aclanthology.org/2023.findings-emnlp.715/}}

@inproceedings{wang2021adamnscl,
  title={Training Networks in Null Space of Feature Covariance for Continual Learning},
  author={Wang, Shipeng and Li, Xiaorong and Sun, Jian and Xu, Zongben},
  booktitle={Proceedings of the IEEE/CVF Conference on Computer Vision and Pattern Recognition},
  pages={184--193},
  year={2021},
  url={https://openaccess.thecvf.com/content/CVPR2021/html/Wang_Training_Networks_in_Null_Space_of_Feature_Covariance_for_Continual_CVPR_2021_paper.html}
}

@article{qwen3_2025,
 title={Qwen3 Technical Report}, author={Yang, An and Li, Anfeng and Yang, Baosong and Zhang, Beichen and Hui, Binyuan and Zheng, Bo and Yu, Bowen and Gao, Chang and Huang, Chengen and Lv, Chenxu and Zheng, Chujie and Liu, Dayiheng and Zhou, Fan and Huang, Fei and Hu, Feng and Ge, Hao and Wei, Haoran and Lin, Huan and Tang, Jialong and Yang, Jian and Tu, Jianhong and Zhang, Jianwei and Yang, Jianxin and Yang, Jiaxi and Zhou, Jing and Zhou, Jingren and Lin, Junyang and Dang, Kai and Bao, Keqin and Yang, Kexin and Yu, Le and Deng, Lianghao and Li, Mei and Xue, Mingfeng and Li, Mingze and Zhang, Pei and Wang, Peng and Zhu, Qin and Men, Rui and Gao, Ruize and Liu, Shixuan and Luo, Shuang and Li, Tianhao and Tang, Tianyi and Yin, Wenbiao and Ren, Xingzhang and Wang, Xinyu and Zhang, Xinyu and Ren, Xuancheng and Fan, Yang and Su, Yang and Zhang, Yichang and Zhang, Yinger and Wan, Yu and Liu, Yuqiong and Wang, Zekun and Cui, Zeyu and Zhang, Zhenru and Zhou, Zhipeng and Qiu, Zihan}, journal={arXiv preprint arXiv:2505.09388}, year={2025}, eprint={2505.09388}, archivePrefix={arXiv}, url={https://arxiv.org/abs/2505.09388}}

@inproceedings{thede2025wikibigedit,
  title={{WikiBigEdit}: Understanding the Limits of Lifelong Knowledge Editing in {LLMs}},
  author={Thede, Lukas and Roth, Karsten and Bethge, Matthias and Akata, Zeynep and Hartvigsen, Thomas},
  booktitle={Proceedings of the 42nd International Conference on Machine Learning},
  series={Proceedings of Machine Learning Research}, volume={267}, pages={59326--59354}, year={2025},
  publisher={PMLR}, url={https://proceedings.mlr.press/v267/thede25a.html}
}

@inproceedings{ammar-khodja-etal-2024-wikifactdiff-large,
  title={{WikiFactDiff}: A Large, Realistic, and Temporally Adaptable Dataset for Atomic Factual Knowledge Update in Causal Language Models},
  author={Ammar Khodja, Hichem and B{\'e}chet, Fr{\'e}d{\'e}ric and Brabant, Quentin and Nasr, Alexis and Lecorv{\'e}, Gw{\'e}nol{\'e}},
  booktitle={Proceedings of the 2024 Joint International Conference on Computational Linguistics, Language Resources and Evaluation (LREC-COLING 2024)},
  pages={17614--17624}, year={2024}, month=may, publisher={ELRA and ICCL},
  doi={10.63317/5kmnn4c4u35s}, url={https://aclanthology.org/2024.lrec-main.1532/}
}

@misc{ticlm2025,
  title = {{TiC-LM}: A Web-Scale Benchmark for Time-Continual {LLM} Pretraining},
  author = {Li, Jeffrey and Armandpour, Mohammadreza and Mirzadeh, Seyed Iman and Mehta, Sachin and Shankar, Vaishaal and Vemulapalli, Raviteja and Bengio, Samy and Tuzel, Oncel and Farajtabar, Mehrdad and Pouransari, Hadi and Faghri, Fartash},
  booktitle = {Proceedings of the 63rd Annual Meeting of the Association for Computational Linguistics (Volume 1: Long Papers)},
  pages = {32231--32273}, year = {2025}, publisher = {Association for Computational Linguistics},
  doi = {10.18653/v1/2025.acl-long.1551}, url = {https://aclanthology.org/2025.acl-long.1551/}
}

@misc{meng2026rethinking,
  title = {Rethinking Transfer in Continual Learning: A Replay-Based Realisation},
  author = {Meng, Yang and Liu, Zhenya and Zhao, Zhuokai and Chen, Yuxin},
  year = {2026}, note = {arXiv:2607.15587}, eprint = {2607.15587}, archiveprefix = {arXiv},
  url = {https://arxiv.org/abs/2607.15587}
}

@misc{locft2026,
  title = {Fine-tuning Done Right in Model Editing},
  author = {Yang, Wanli and Tang, Rui and Zang, Hongyu and Su, Du and Cao, Qi and Wang, Jingang and Shen, Huawei and Cheng, Xueqi and Sun, Fei},
  year = {2026}, note = {ICLR 2026; arXiv:2509.22072}, eprint = {2509.22072}, archiveprefix = {arXiv}, doi = {10.48550/arXiv.2509.22072}, url = {https://arxiv.org/abs/2509.22072}
}

@misc{nas2026,
  title = {Norm Anchors Make Model Edits Last},
  author = {Liu, Mingda and Zhu, Zhenghan and Miao, Ze'an and Fujisawa, Katsuki},
  year = {2026}, note = {arXiv:2602.02543}, eprint = {2602.02543}, archiveprefix = {arXiv}, doi = {10.48550/arXiv.2602.02543}, url = {https://arxiv.org/abs/2602.02543}
}

@misc{crispedit2026,
  title = {CrispEdit: Low-Curvature Projections for Scalable Non-Destructive {LLM} Editing},
  author = {Ikram, Zarif and Firouzkouhi, Arad and Tu, Stephen and Soltanolkotabi, Mahdi and Rashidinejad, Paria},
  year = {2026}, note = {ICML 2026; arXiv:2602.15823}, eprint = {2602.15823}, archiveprefix = {arXiv}, doi = {10.48550/arXiv.2602.15823}, url = {https://arxiv.org/abs/2602.15823}
}

@misc{stableedit2026,
  title = {More Edits, More Stable: Understanding the Lifelong Normalization in Sequential Model Editing},
  author = {Ma, Xin and Chen, Wei and Liu, Qi and Xu, Derong and Zheng, Zhi and Xu, Tong and Chen, Enhong},
  year = {2026}, note = {ICML 2026; arXiv:2605.11836}, eprint = {2605.11836}, archiveprefix = {arXiv}, doi = {10.48550/arXiv.2605.11836}, url = {https://arxiv.org/abs/2605.11836}
}

@misc{ultraedit2026,
  title = {UltraEdit: Training-, Subject-, and Memory-Free Lifelong Editing in Language Models},
  author = {Gu, Xiaojie and Huang, Ziying and Gu, Jia-Chen and Zhang, Kai},
  year = {2026}, note = {TMLR 2026; arXiv:2505.14679}, eprint = {2505.14679}, archiveprefix = {arXiv}, doi = {10.48550/arXiv.2505.14679}, url = {https://arxiv.org/abs/2505.14679}
}

@inproceedings{fan2018learning,
  title = {Learning to Teach},
  author = {Fan, Yang and Tian, Fei and Qin, Tao and Li, Xiang-Yang and Liu, Tie-Yan},
  booktitle = {International Conference on Learning Representations},
  year = {2018}
}

@inproceedings{graves2017automated,
  title = {Automated Curriculum Learning for Neural Networks},
  author = {Graves, Alex and Bellemare, Marc G. and Menick, Jacob and Munos, R{\'e}mi and Kavukcuoglu, Koray},
  booktitle = {Proceedings of the 34th International Conference on Machine Learning},
  series = {Proceedings of Machine Learning Research},
  volume = {70},
  pages = {1311--1320},
  year = {2017},
  publisher = {PMLR},
  url = {https://proceedings.mlr.press/v70/graves17a.html}
}

\appendix
\section{Additional 150K WikiBigEdit Evaluation}
\label{app:wikibigedit-150k}

We compare \method{} (Full FT) with Full-history Replay (LoRA) at 150K using the same Match definitions as the 100K evaluation. \method{} exceeds Full-history Replay on Original, Rephrase, Persona, and Locality Match by 3.78, 9.06, 22.28, and 7.42 percentage points, respectively (Table~\ref{tab:wikibigedit-150k}).

\begin{table}[h]
\centering
\small
\setlength{\tabcolsep}{3.0pt}
\renewcommand{\arraystretch}{1.12}
\caption{WikiBigEdit Match accuracy (\%) after 150,000 continual updates. The best result in each column is shown in bold.}
\label{tab:wikibigedit-150k}
\resizebox{\columnwidth}{!}{%
\begin{tabular}{@{}lrrrr@{}}
\toprule
Method & Original & Rephrase & Persona & Locality \\
\midrule
Full-history Replay (LoRA) & 93.75 & 78.71 & 45.31 & 22.66 \\
\textbf{\method{} (Full FT, ours)} & \textbf{97.53} & \textbf{87.77} & \textbf{67.59} & \textbf{30.08} \\
\bottomrule
\end{tabular}%
}
\end{table}

\section{WikiBigEdit Replay Volume}
\label{app:replay-volume}

Table~\ref{tab:wikibigedit-replay-volume} reports cumulative fact presentations over 100K WikiBigEdit facts. One fact presentation counts one selected fact once; total fact presentations include the current facts and historical replay. \method{} and Random Replay use full-parameter updates, whereas Full-history Replay uses LoRA. Full-history Replay trains new facts in blocks of up to 375; a block also ends at each native-interval boundary (every 12.5K facts) and at each evaluation checkpoint. At the start of every block, it replays all previously observed facts. Its cumulative historical replay count is $\sum_b s_b$, where $s_b$ is the number of previously observed facts at the start of block $b$.

\begin{table}[t]
\centering
\scriptsize
\setlength{\tabcolsep}{4pt}
\renewcommand{\arraystretch}{1.12}
\caption{Cumulative WikiBigEdit fact presentations, including current facts and historical replay.}
\label{tab:wikibigedit-replay-volume}
\resizebox{\columnwidth}{!}{%
\begin{tabular}{@{}lrrrr@{}}
\toprule
Method & 3K & 10K & 50K & 100K \\
\midrule
\multicolumn{5}{@{}l}{\textit{Fact presentations (current + replay)}} \\
\method{} (Full FT) & 14,801 & 49,516 & 246,364 & 494,424 \\
Random Replay (Full FT) & 14,984 & 49,984 & 249,984 & 499,984 \\
Full-history Replay (LoRA) & 13,500 & 141,625 & 3,440,625 & 13,682,125 \\
\bottomrule
\end{tabular}
}
\end{table}

\section{Writer Action Utilization}
\label{app:writer-action-utilization}



\begin{figure*}[!t]
\centering
\includegraphics[width=\textwidth]{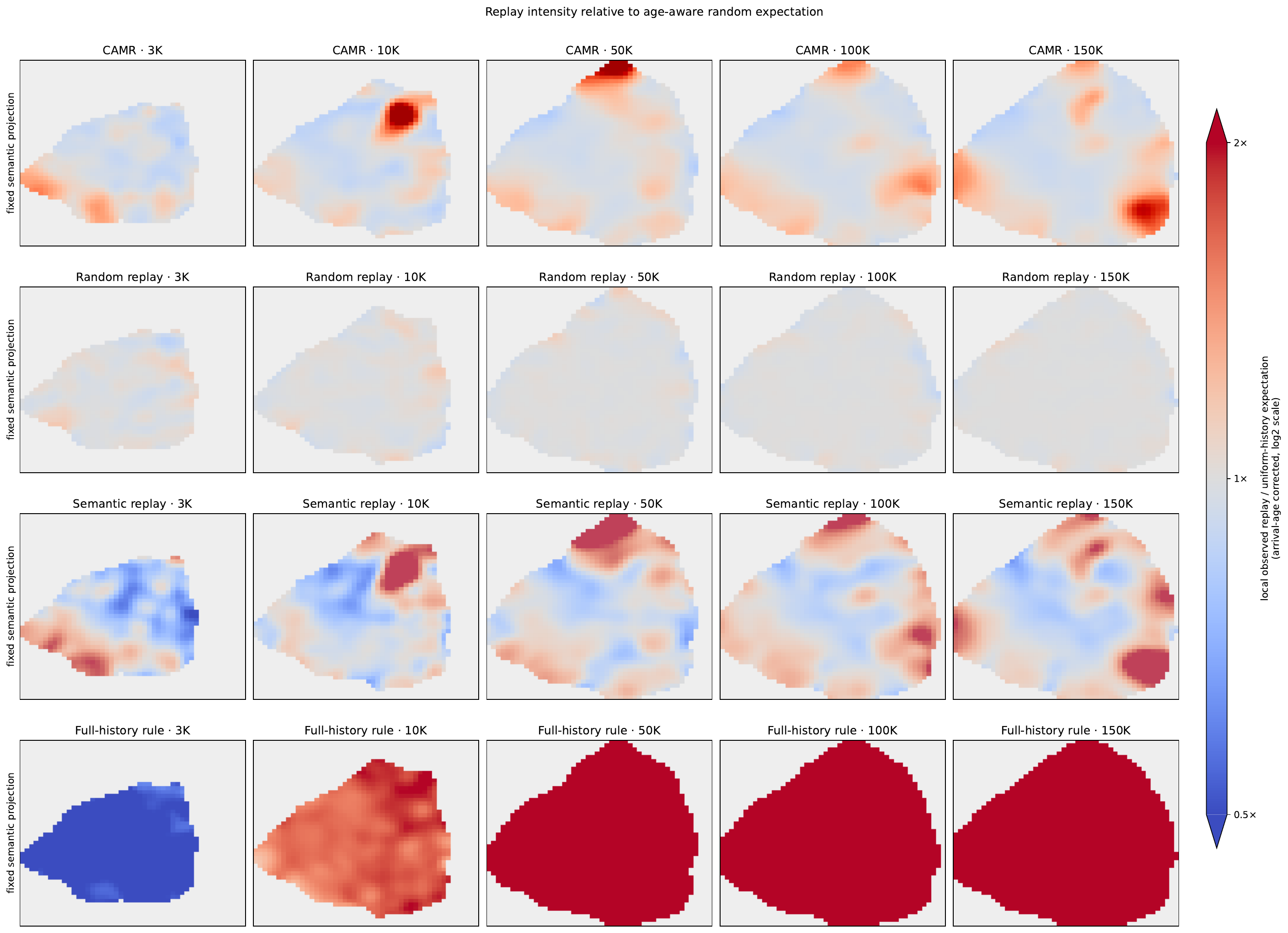}
\caption{Age- and eligibility-corrected semantic replay allocation across the WikiBigEdit stream. Rows show \method{}, count-matched Random Replay from \method{}'s logged eligible history pools, count-matched cosine Top-8 Semantic Replay from the same pools, and the theoretical 375-record Full-history Replay at 3K, 10K, 50K, 100K, and 150K. Each panel reports local cumulative replay intensity divided by the exact event-wise Random Replay expectation after identical fixed-grid smoothing. Neutral gray denotes $1\!\times$ expectation, red denotes over-allocation, and blue denotes under-allocation on the shared log$_2$ scale. All checkpoints use the same frozen representation and PCA projection.}
\label{fig:semantic-replay-density-ratio}
\end{figure*}

\begin{table}[t]
\centering
\small
\caption{Cumulative writer-action shares (\%) under the \ignoreact{} gate.}
\label{tab:wikibigedit-action-ratios}
\begin{tabular*}{\columnwidth}{@{\extracolsep{\fill}}lrrrr@{}}
\toprule
Stream & Size & \addact{} & \mergeact{} & \ignoreact{} \\
\midrule
WikiBigEdit & 3K   & 67.83 & 32.17 & 0.00 \\
            & 10K  & 59.55 & 40.45 & 0.00 \\
            & 50K  & 71.75 & 28.25 & 0.00 \\
            & 100K & 69.73 & 30.27 & 0.00 \\
\midrule
APPS & 5K & 58.94 & 21.92 & 19.14 \\
\bottomrule
\end{tabular*}
\end{table}

Table~\ref{tab:wikibigedit-action-ratios} reports a controlled
action audit using the \ignoreact{} gate defined in Section~\ref{sec:policy}
from the original experiments.
Under this gate, \ignoreact{} requires an equivalent active
record, current NLL $\leq\tau_m$, no retirement triggered
by the arrival, and a fresh greedy answer matching
the full target.
No arrival in the 100K WikiBigEdit stream satisfied
the duplicate/equivalence prerequisite, so \ignoreact{}
was never selected.

For each task, we construct a 5K-arrival stream by randomly
subsampling larger training sets or supplementing smaller ones
with randomly sampled training examples. All methods use the same
resulting stream and order.

We used the same gate in a separate diagnostic with
repeated APPS examples.
Starting from the original single-pass stream of
2,639 examples, we appended 2,361 exact input--output
duplicates of previously observed active examples,
producing 5,000 arrivals.
Over this stream, the writer selected \addact{} 2,947 times
(58.94\%), \mergeact{} 1,096 times (21.92\%), and
\ignoreact{} 957 times (19.14\%).
This diagnostic shows that, under repeated exposure,
the writer selectively skips direct supervision
of examples that satisfy the mastery criteria.


\begin{figure*}[!t]
\centering
\includegraphics[width=\textwidth]{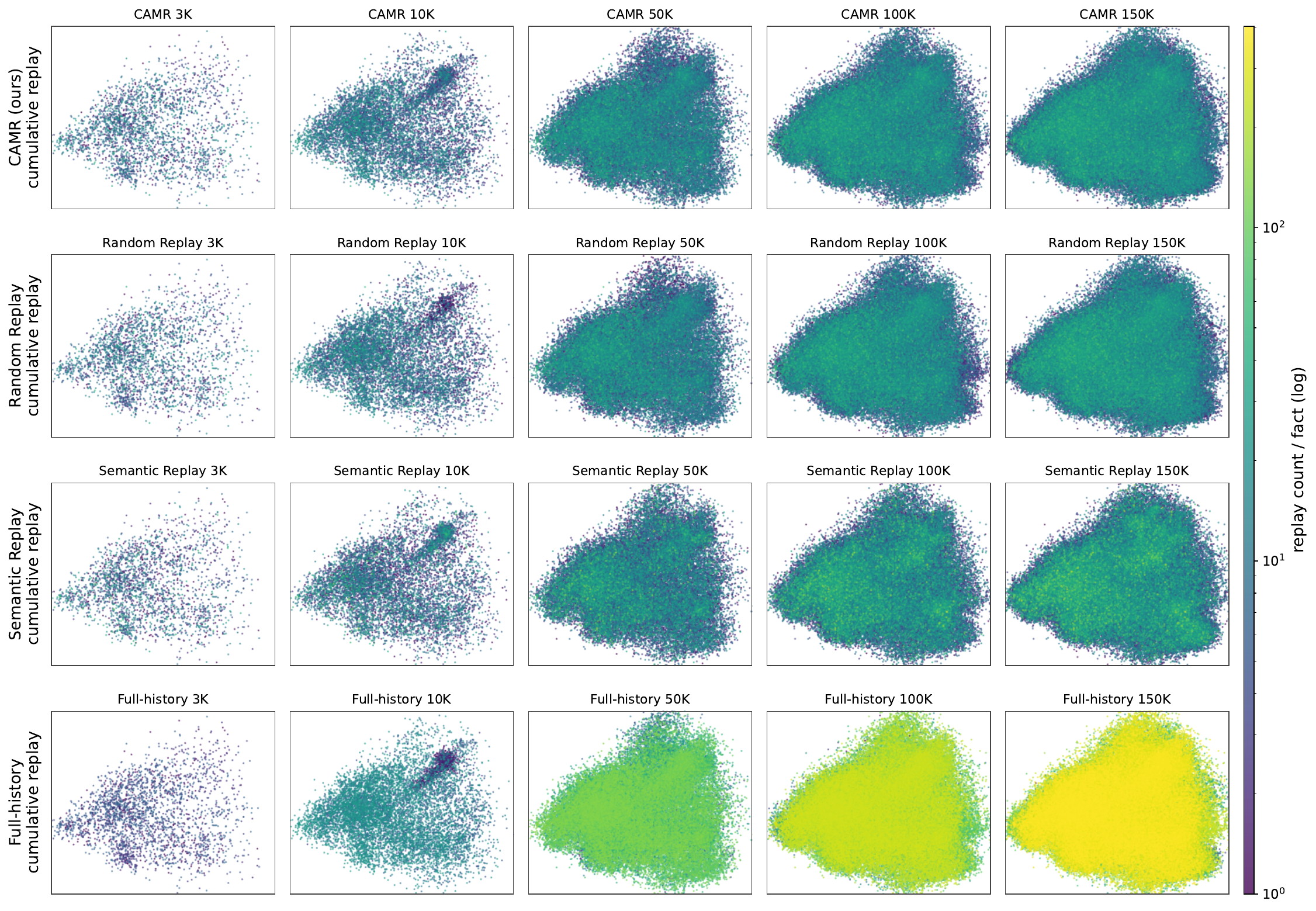}
\caption{Unnormalized cumulative replay counts in the same semantic space. Rows and checkpoints follow Figure~\ref{fig:semantic-replay-density-ratio}. Each point is a replayed fact at its actual PCA coordinate, and color gives its cumulative replay count on a shared logarithmic scale. The shared axes, frozen representation, and PCA projection make semantic coverage and per-fact repetition directly comparable across methods and stream prefixes.}
\label{fig:semantic-replay-raw-counts}
\end{figure*}

\begin{table}[h]
\centering
\caption{Complete set of \method{} hyperparameters governing memory,
replay, the writer architecture, and the learning objectives.}
\label{tab:camr-hyperparameters}
\footnotesize
\setlength{\tabcolsep}{4pt}
\renewcommand{\arraystretch}{1.02}
\begin{tabularx}{\columnwidth}{@{}X>{\raggedleft\arraybackslash}p{0.42\columnwidth}@{}}
\toprule
Hyperparameter & Value \\
\midrule
\multicolumn{2}{@{}c}{\textit{Training and optimization}} \\ \midrule
Base language model & Qwen3-8B-Base \\
Language-model update & Full-parameter SFT \\
Language-model optimizer & AdamW \\
Language-model learning rate & $2\times10^{-6}$ \\
Writer optimizer & Adam \\
Writer learning rate & $3\times10^{-4}$ \\
Optimizer steps per arrival & 1 \\
Logical batch size per arrival & $1+|\mathcal{R}_t|\leq 5$ facts \\
Weight decay / gradient clip & $0$ / $1.0$ \\
\addlinespace[2pt]
\midrule
\multicolumn{2}{@{}c}{\textit{Memory and replay}} \\ \midrule
Probe cap $P$ & 4 \\
Retrieved-candidate cap & 64 \\
Candidate similarity threshold $\tau_s$ & 0.65 \\
Mastery NLL threshold $\tau_m$ & 0.25 \\
Per-arrival replay cap $K$ & 4 \\
\mergeact{} uniform-replay quota & 2 of 4 slots \\
\mergeact{} candidate quota & Up to 2 of 4 slots \\
\mergeact{} candidate selection & Writer-selected from $\mathcal{C}_t$, without replacement \\
\addlinespace[2pt]
\midrule
\multicolumn{2}{@{}c}{\textit{Writer and learning objectives}} \\ \midrule
Writer projection dimension $d_h$ & 32 \\
Writer rollout length $|\mathcal{B}|$ & 4 events \\
Current-example loss weight $\alpha$ & 0.5 \\
Retention reward weight $\lambda_h$ & 1.0 \\
Cost penalty weight $\lambda_c$ & 0.01 \\
Cost reference $C_{\mathrm{ref}}$ & 128 tokens if $L_{\max}<128$; 1,024 tokens otherwise \\
Discount factor $\gamma$ & 0.95 \\
Entropy coefficient $\beta$ & 0.01 \\
Critic-loss coefficient $c_v$ & 0.5 \\
\bottomrule
\end{tabularx}
\par\vspace{2pt}
{\scriptsize
\setlength{\tabcolsep}{2pt}
\begin{tabular*}{\columnwidth}{@{\extracolsep{\fill}}lrr@{}}
\multicolumn{3}{c}{\textit{Training arrivals and train/output token caps}} \\ \midrule
\textit{Dataset} & \textit{Arrivals} & \textit{Train/output caps} \\
WikiBigEdit & 100,000 & 512 / 64 \\
WikiFactDiff & 36,000 & 256 / 16 \\
TRACE & 5,000/task & 9,216 / 1,024 \\
Cross-domain & 5,000/task & 16,384 / 4,096 \\
FOMC/NumGLUE & 5,000/task & 640 / 8--128 \\
Py150 & 5,000 & 4,608 / 512 \\
\bottomrule
\end{tabular*}
}
\end{table}

\section{Complete CAMR Hyperparameter Specification}
\label{app:objective-hyperparameters}

Table~\ref{tab:camr-hyperparameters} reports the complete set of
hyperparameters introduced in the \method{} formulation, together
with their values in the reference configuration.

These values define the reference \method{} configuration.
The choices $P=4$ and $K=4$ bound the per-arrival probe
and replay loads, while $\tau_s=0.65$ and $\tau_m=0.25$ impose
selective compatibility and mastery criteria.
The 32-dimensional projection keeps the writer compact.
Setting $\alpha=0.5$ gives equal weight to the current-example
term and the aggregate replay term in Equation~\ref{eq:lm_loss}.
The reward weights retain the historical-gain term at
unit scale and apply a token-cost penalty in
Equation~\ref{eq:reward}, normalized by the dataset-specific
$C_{\mathrm{ref}}$ values in Table~\ref{tab:camr-hyperparameters}.
Specifically, letting $L_{\max}$ denote the maximum sequence length
of a dataset, we set $C_{\mathrm{ref}}=128$ when $L_{\max}<128$
and $C_{\mathrm{ref}}=1024$ otherwise.
Finally, $\gamma=0.95$ preserves credit within the four-event
rollout, $\beta=0.01$ supplies a modest entropy bonus, and
$c_v=0.5$ gives the critic loss a moderate coefficient in
Equations~\ref{eq:writer_returns} and~\ref{eq:writer_loss}.

\section{Semantic Replay Trajectory Analysis}
\label{app:semantic-trajectory}

We compare the cumulative replay allocations of \method{}, Random Replay, Semantic Replay, and the theoretical Full-history Replay at 3K, 10K, 50K, 100K, and 150K. All facts are projected from the frozen 768-dimensional memory representation. PCA is fitted once on all 150K facts and shared across checkpoints; the first two components explain 8.61\% of the variance. Random Replay and Semantic Replay match \method{}'s per-event replay budget over the same logged safe, non-probe history pools; the former samples uniformly without replacement, whereas the latter selects up to eight nearest facts by cosine similarity. The full-history allocation follows the 375-record block rule described in Appendix~\ref{app:replay-volume}.

Figure~\ref{fig:semantic-replay-density-ratio} removes the cumulative advantage of earlier-arriving facts and the event-wise variation in eligible history. Random Replay remains close to the neutral $1\!\times$ level across regions and checkpoints. \method{} forms localized regions of elevated and reduced intensity as the stream grows, showing targeted semantic reweighting within a bounded replay budget. Semantic Replay follows nearest-neighbor structure and produces stronger hotspots and cold regions. The Full-history Replay is below the matched-budget expectation at 3K and reaches the upper range across the map from 50K onward as repeated full-history sweeps accumulate.

The raw-count view in Figure~\ref{fig:semantic-replay-raw-counts} shows both semantic coverage and repeated selection. Under the matched budget, \method{} maintains broad support while concentrating additional presentations in selected neighborhoods; Random Replay distributes presentations diffusely, whereas Semantic Replay repeats a smaller set of nearest neighbors more often. At 100K, the three strategies have replayed 90.4\%, 88.8\%, and 83.9\% of the arrived facts, respectively. Full-history Replay reaches 99.8\% coverage but produces a broadly elevated count field as its cumulative replay volume grows. Together, the normalized and raw-count views show that \method{} combines broad semantic support with selective allocation of a bounded replay budget.

\section{Core-knowledge replay under stream growth.}
Table~\ref{tab:wikibigedit-core3k-replay-share} reports the share of replay slots assigned to a fixed high-frequency core-3K set at each WikiBigEdit checkpoint.

\begin{table}[h]
\centering
\small
\setlength{\tabcolsep}{4pt}
\renewcommand{\arraystretch}{1.12}
\caption{Share (\%) of replay slots allocated to the fixed high-frequency core-3K set as the WikiBigEdit history grows.}
\label{tab:wikibigedit-core3k-replay-share}
\begin{tabular}{@{}lrrrr@{}}
\toprule
Method & 3K & 10K & 50K & 100K \\
\midrule
\method{} & \textbf{25.54} & \textbf{20.93} & \textbf{9.74} & \textbf{7.30} \\
Random Replay & 19.49 & 13.56 & 5.94 & 3.02 \\
\bottomrule
\end{tabular}
\end{table}

Both methods devote a smaller share of replay to the core-3K set as the history expands. Random Replay samples uniformly from the available history, so its core share falls from 19.49\% at 3K to 3.02\% at 100K. When \mergeact{} is selected, \method{} reserves part of its replay budget for compatible retrieved facts, allowing core facts related to later arrivals to receive targeted rehearsal. Its core share remains higher at every checkpoint, decreasing from 25.54\% to 7.30\%. At 100K, \method{} assigns 2.42$\times$ the replay share of Random Replay to the core-3K set, reducing the dilution of core knowledge as the replay pool grows.

\FloatBarrier

\clearpage

\clearpage

\end{document}